\documentclass[runningheads]{llncs}

\usepackage{eccv}
\usepackage{eccvabbrv}

\usepackage{graphicx}
\usepackage{amsmath,amssymb}
\usepackage{booktabs}

\usepackage{multirow}
\usepackage{xcolor}
\usepackage{colortbl}
\usepackage{makecell}
\usepackage{array}
\usepackage{wrapfig}
\usepackage{subcaption}

\usepackage[accsupp]{axessibility} 
\usepackage{url}
\usepackage[breaklinks,colorlinks,citecolor=eccvblue]{hyperref}

\usepackage{orcidlink}

\begin{document}
\title{SPECSIA: Stylization Dataset for Novel-View Enhancement in Drawing-based 3D Animation} 

\titlerunning{SPECSIA}

\author{
Kyuwon Kim\inst{1}\orcidlink{0009-0004-2013-0467} \and
Sunjae Yoon\inst{2}\orcidlink{0000-0001-7458-5273} \and
Chang D. Yoo\thanks{Corresponding author.}\inst{1}\orcidlink{0000-0002-0756-7179}
}

\authorrunning{K.~Kim et al.}

\institute{
\textsuperscript{1}\,School of Electrical Engineering, KAIST, Daejeon, Republic of Korea
\quad
\textsuperscript{2}\,Department of AI, Chung-Ang University, Seoul, Republic of Korea\\
\textsuperscript{1}\,\email{\{rbdnjs7830, cd\_yoo\}@kaist.ac.kr}
\quad
\textsuperscript{2}\,\email{sunjaeyoon@cau.ac.kr}
}

\maketitle

\begin{figure}
  \centering
  \includegraphics[width=0.90\textwidth]{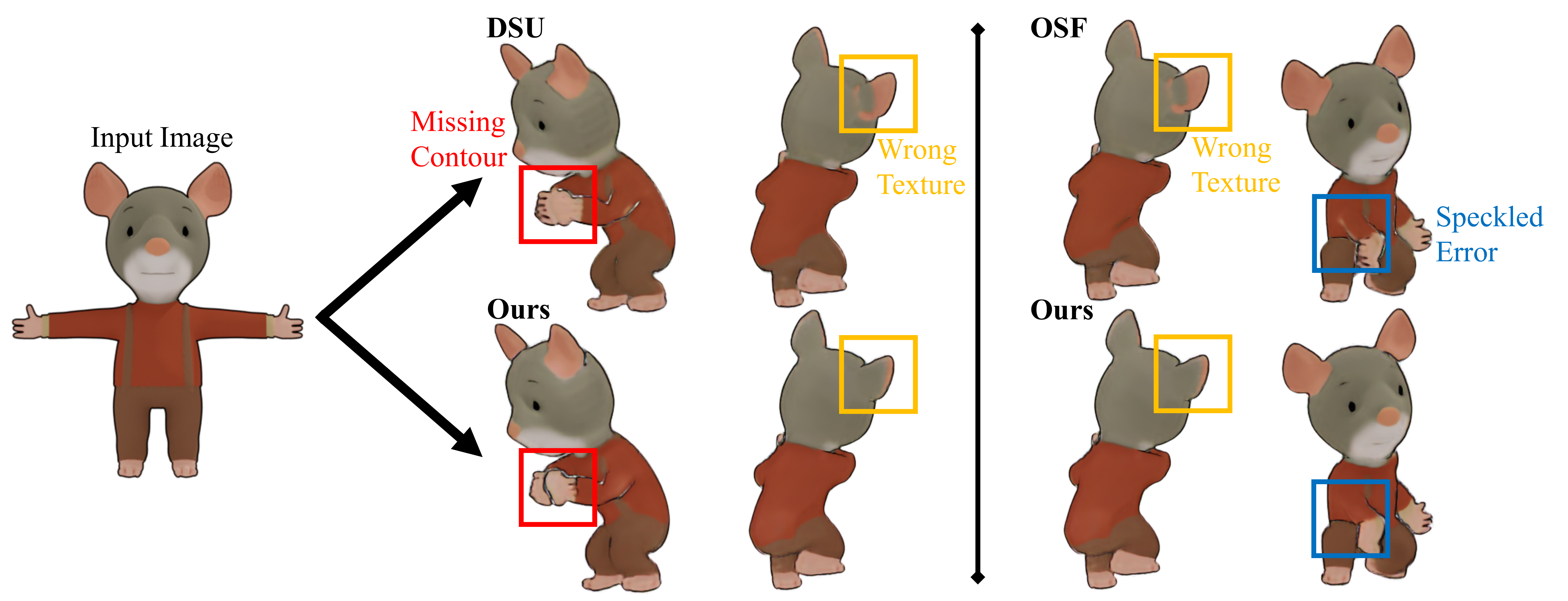}
  \caption{Comparative animation results using DraViE (ours), DrawingSpinUp (DSU) \cite{zhou2024drawingspinup}, and the Occlusion-robust Stylization Framework (OSF) \cite{yoon2025occlusion}. DraViE reduces novel-view artifacts while preserving the input drawing style.
  }
  \label{fig:teaser}
\end{figure}

\begin{abstract}
    Generating animation from a single 2D drawing is challenging because the output must preserve character appearance while remaining plausible and temporally coherent under motion.
    Existing drawing-based 3D animation pipelines often use sample-wise 2D refinement to align animated renderings with the input image, but such optimization tends to overfit to the observed view and fails to correct projection-induced artifacts in novel views.
    To address this limitation, we introduce SPECSIA-15K, a paired stylization dataset containing 14,980 artifact-corrupted projection/refinement-target pairs from 1,498 3DBiCar characters.
    We further present DraViE (Drawing-based View Enhancement), a lightweight plug-and-play module trained with data-level priors to remove novel-view artifacts while preserving style and motion plausibility.
    Experiments show consistent gains in novel-view fidelity and temporal coherence with lower per-character adaptation cost than sample-wise fine-tuning.
    Project page: \url{https://rbdnjs7830.github.io/SPECSIA/}
\end{abstract}

\begin{figure}[t]
  \centering
  \begin{subfigure}{0.65\textwidth}
    \includegraphics[width=1.0\textwidth]{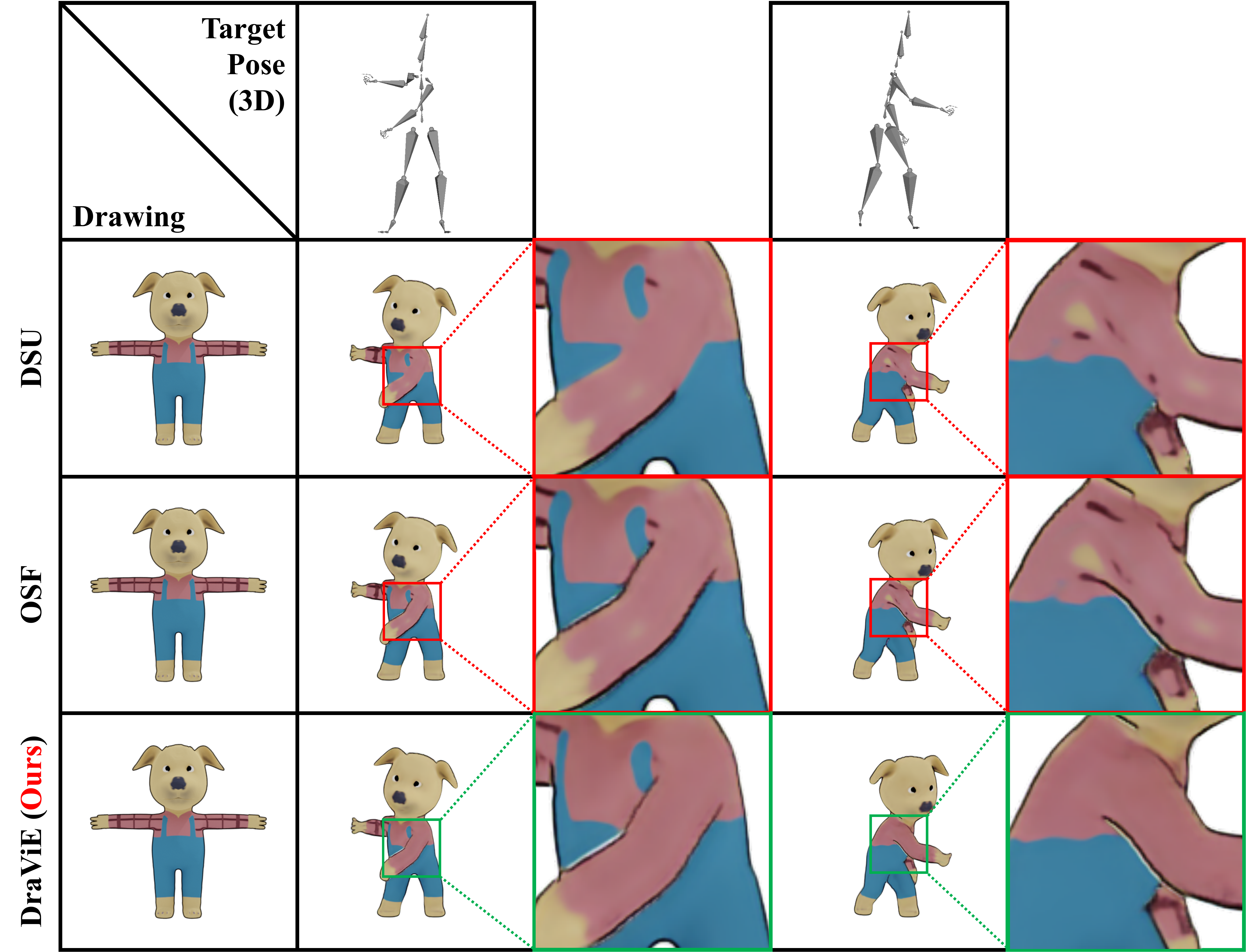}
    \caption{Qualitative analysis on novel-view}
  \end{subfigure}
  \begin{subfigure}{0.33\textwidth}
    \includegraphics[width=1.0\textwidth]{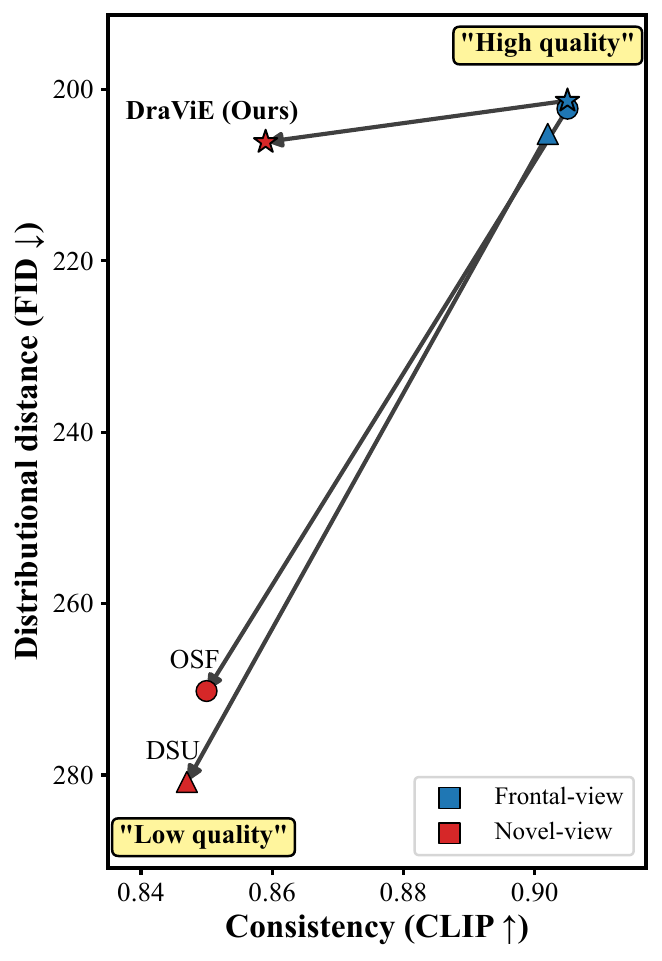}
    \caption{Quality change under viewpoint variation}
  \end{subfigure}
  \caption{Quality degradation in novel views.
  (a) Existing pipelines produce artifacts in novel-view renderings, such as missing contours and speckle-like regions (zoomed in red boxes).
  DSU: DrawingSpinUp~\cite{zhou2024drawingspinup}, OSF: Occlusion-robust Stylization Framework~\cite{yoon2025occlusion}.
  (b) Joint plot of CLIP (semantic consistency)~\cite{pmlr-v139-radford21a} and FID (distributional fidelity)~\cite{NIPS2017_8a1d6947} for frontal vs. novel views.
  Triangles/circles indicate DSU/OSF, and blue/red points indicate frontal/novel views.
  Lines connect the two for each method, highlighting the degradation in novel views.
  The plot follows the same evaluation setting as Table~\ref{tab:quantitative}.}
  \label{fig:intro1}
\end{figure}

\begin{figure}[t]
  \centering
  \includegraphics[width=1.0\textwidth]{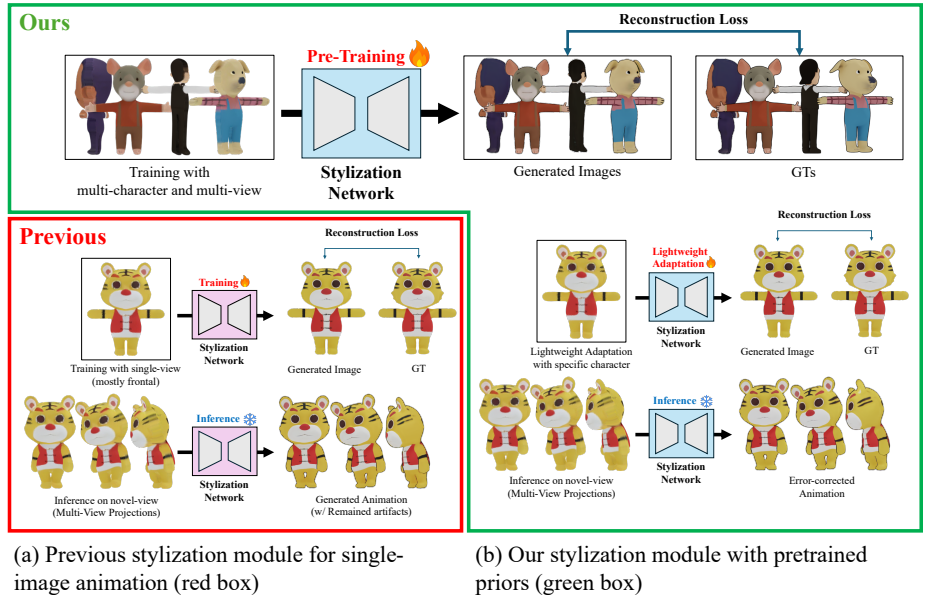}
  \caption{Comparison between the previous stylization pipeline and our approach. (a) Illustration of the previous stylization network with sample-wise alignment. Because sample-wise alignment is tied to the input pose, the module may fail to correct errors that emerge in 3D rendering or 2D projection.
  (b) Our approach: stylization network with a data-level pre-trained prior.}
  \label{fig:methods}
\end{figure}

\section{Introduction}
\label{sec:intro}

Single-image animation \cite{Siarohin_2019_NeurIPS, hu2023animateanyone, zhou2024drawingspinup, yoon2025occlusion} has become increasingly important for applications such as character animation, content creation, and virtual avatars. However, generating time-varying motion from a single input image remains challenging: the output should preserve the input character's visual identity (e.g., color, texture, and silhouette) while maintaining physically plausible motion and temporal coherence \cite{xu2023magicanimate, guo2023animatediff, zhou2024drawingspinup, yoon2025occlusion}. This challenge is further amplified because the input is typically limited to a single view (typically a frontal view), making it difficult to maintain stable appearance in unseen or novel views. Figs.~\ref{fig:teaser} and~\ref{fig:intro1} show representative results and the motivating novel-view degradation.

Many recent approaches adopt 3D-assisted pipelines that reconstruct a 3D structure (e.g., a mesh) from the input, apply target motion in 3D, and render the animated result back to 2D \cite{tang2023dreamgaussian, long2023wonder3d}. However, correcting errors in 3D is expensive and tightly coupled to the reconstruction pipeline. As a practical alternative, drawing-based animation systems rely on sample-wise 2D refinement by aligning the input image with 2D projections of the rendered animation \cite{zhou2024drawingspinup, yoon2025occlusion}. While effective at improving appearance in the observed view, such sample-wise optimization often overfits to the input view and degrades quality in unseen views, because alignment objectives enforce pixel-/feature-level consistency and cannot reliably correct artifacts originating from 3D rendering and projection.

A key bottleneck, therefore, is the lack of scalable supervision for improving unseen-view quality under large viewpoint changes and view-dependent projection artifacts. This scarcity makes it difficult to learn correction priors that generalize across characters, motions, and viewpoints. To address this gap, we introduce \textbf{SPECSIA-15K}, a stylization dataset designed for novel-view enhancement in drawing-based 3D animation. SPECSIA-15K is constructed from multi-view renderings of 3DBiCar~\cite{luo2023rabit}: we sample 10 views per object and form 14,980 paired examples of artifact-corrupted 2D projections and corresponding targets for refinement. This paired supervision enables data-level learning of correction priors that generalize across diverse characters, motions, and viewpoints.

Leveraging SPECSIA-15K, we present \textbf{DraViE} (Drawing-based View Enhancement), a lightweight pre-trained module that performs plug-and-play post-correction on rendered outputs. Trained at the data level rather than via sample-wise optimization, DraViE learns priors that correct novel-view artifacts while preserving drawing style and motion plausibility.
Finally, while data-level priors improve generalization, test-time inputs may still exhibit unique drawing styles (e.g., line thickness or color palette). To handle this in a practical and low-cost manner, we optionally perform a lightweight adaptation on the single reference drawing using a small number of updates. This improves alignment to the specific input style while preserving the learned priors and avoiding strong frontal-view overfitting.
As a result, DraViE enables view-consistent, high-quality animation with lower per-character adaptation cost in our setting, while remaining easy to integrate into diverse motion generation pipelines as a simple post-processing component.

Overall, this paper reveals the limitations of sample-wise alignment in existing drawing-based 3D animation pipelines, which often overfit to the observed view.
To enable scalable supervision for novel-view enhancement, we introduce SPECSIA-15K, a scalable stylization dataset that supports learning and evaluation.
Building on this dataset, we propose DraViE, a data-level pre-trained, plug-and-play module for novel-view error correction, yielding improvements in novel-view fidelity and temporal coherence across motions and viewpoints.

\section{Related Works}
\label{sec:related_works}

\subsection{Image-to-3D Animation \& Hand-drawn Animation}
Prior work on single-image animation combines deformation modeling with image synthesis to transfer motion while preserving appearance \cite{aberman2019learning, albahar2021pose}.
Recent works based on editing- and mesh-guided methods further improve controllability for non-rigid editing and structure-aware manipulation \cite{10.1145/1057432.1057456, koo2024flexiedit, koo2025flowdrag}.
Nevertheless, under single-view inputs, depth ambiguity and occlusion make it difficult to maintain identity and temporal coherence, especially under viewpoint changes \cite{Siarohin_2019_NeurIPS, xu2023magicanimate, yoon2024tpc}.
These limitations have motivated 3D-assisted pipelines that reconstruct and animate a 3D representation, often aided by multi-view generation to improve reconstruction quality \cite{tang2023dreamgaussian, long2023wonder3d, liu2023syncdreamer}.
However, for hand-drawn characters, rendering and projection can distort lines and textures, and existing pipelines commonly rely on sample-wise optimization to match the input style \cite{zhou2024drawingspinup, yoon2025occlusion}.
In this context, the scarcity of scalable supervision for improving unseen-view quality remains a key challenge for learning view-consistent refinement.
This motivates data-level resources and learning objectives that explicitly provide scaled supervision for improving novel-view quality in drawing-based 3D animation.

\subsection{Image Stylization}
Image stylization aims to transform "style" factors (e.g., texture and color) while preserving the content and identity of the input.
Early work formulated stylization via content-style decomposition \cite{gatys2015neuralalgorithmartisticstyle}, and later studies adopted GAN-based image-to-image translation for broader domain transfer \cite{isola2017image, CycleGAN2017}.
More recently, diffusion models have extended stylization beyond pure style change toward quality enhancement and error correction by leveraging generative priors under diverse degradations \cite{10.5555/3495724.3496298, conf/iclr/SongME21, pmlr-v139-nichol21a, 9887996, Choi_2021_ICCV}.
In video and animation, applying stylization independently to each frame often breaks temporal coherence, motivating approaches that enforce temporal consistency for coherent stylization, generation, and diffusion video editing \cite{Ruder_2016, conf/accv/GaoGZY18, guo2023animatediff, yoon2024frag}.
This issue is particularly pronounced for hand-drawn character animation, where style is strongly defined by lines and contours.
Recent pipelines take 3D-animated renderings and projected 2D images as input and apply stylization networks to refine appearance and match the intended drawing style \cite{zhou2024drawingspinup, yoon2025occlusion}.
However, these methods focus on local refinement and may have limited capacity to correct view-dependent rendering/projection errors that emerge in novel views.
Therefore, a stylization formulation that incorporates priors for error correction supported by scaled data across viewpoints is crucial for robust novel-view enhancement in drawing-based animation.

\begin{figure}
  \centering
  \includegraphics[width=1.0\textwidth]{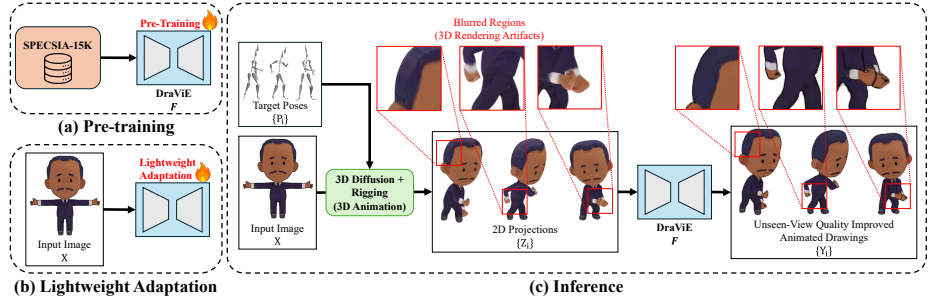}
  \caption{DraViE: Drawing-based View Enhancement. (a) We pretrain DraViE on SPECSIA-15K. (b) We then optionally perform lightweight adaptation to the input character. (c) Given an input drawing and a target motion, a 3D animation system (e.g., 3D diffusion \cite{long2023wonder3d} and rigging \cite{10.1145/1276377.1276467}) produces novel-view projections that often contain view-dependent projection-induced artifacts. DraViE applies prior-guided 2D refinement as a plug-and-play module, reducing view-dependent degradations while preserving the original drawing style. Red boxes highlight artifacts corrected by DraViE.
  }
  \label{fig:pipeline}
\end{figure}

\section{Method}
We address novel-view degradation and artifacts in drawing-based 3D animation pipelines by introducing (i) \textbf{SPECSIA-15K}, a paired stylization dataset for projection refinement, and (ii) \textbf{DraViE} (Drawing-based View Enhancement), a lightweight pre-trained post-correction module. Our goal is not to replace upstream 3D animation pipelines, but to provide scalable supervision and a modular 2D refinement prior for reducing view-dependent projection-level artifacts. Figs.~\ref{fig:methods} and~\ref{fig:pipeline} summarize the workflow and comparison to sample-wise stylization.

A typical drawing-based animation pipeline first reconstructs an animatable 3D character from a single input drawing $X$ using an image-to-3D model (e.g., diffusion-based reconstruction \cite{long2023wonder3d}), applies rigging to enable pose-driven articulation \cite{10.1145/1276377.1276467}, and renders animated frames under a motion sequence $\{P_i\}$.
Here, $i \in \{1,\dots,T\}$ indexes the animation frame (time step), and $P_i$ denotes the target pose at frame $i$.
Given a target pose $P_i$, the system produces a rendered 2D projection $Z_i$ from the user's viewpoint.
In practice, $Z_i$ often exhibits line/contour distortions, texture degradation, and view-dependent artifacts, particularly in unseen viewpoints.
While correcting these errors directly in 3D is expensive and tightly coupled to the reconstruction pipeline, many systems employ 2D refinement modules on projected images for efficiency and modularity.
However, existing approaches are optimized in a sample-wise manner (often per character), which can overfit to the observed view and fail to generalize to novel views.

DraViE performs 2D post-correction on projected frames using an encoder-decoder network $F$ (e.g., U-Net \cite{10.1007/978-3-319-24574-4_28}).
Following prior works that use spatial conditioning to resolve local ambiguities \cite{zhou2024drawingspinup,10.1145/3306346.3323006}, we incorporate a positional hint $Z^{pos}$ as an input to the refinement network. Given the projection $Z_i$ and auxiliary signals (mask and positional hint), DraViE outputs a refined frame:
\begin{equation}
Y_i = F\!\left(Z_i, Z^{mask}_i, Z^{pos}_i\right),
\end{equation}
where $Y_i$ corrects artifacts in $Z_i$ while preserving the drawing style and motion plausibility.
This enables refinement without modifying the upstream pipeline.

\begin{figure}[t]
  \centering
  \includegraphics[width=0.90\textwidth]{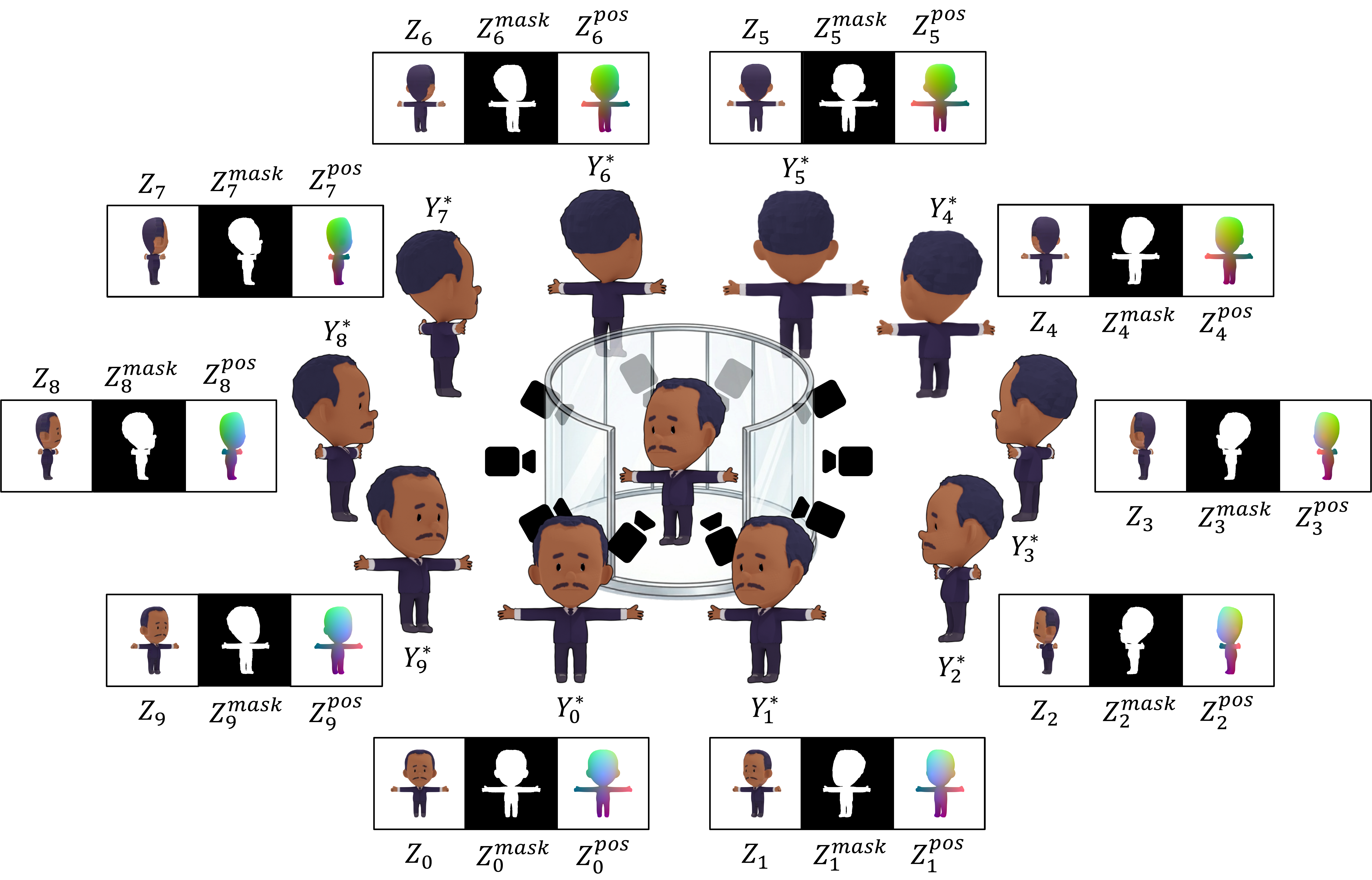}
  \caption{SPECSIA-15K Dataset Construction. We construct paired viewpoint-aligned training data from 1,498 3DBiCar~\cite{luo2023rabit} characters by rendering 10 viewpoints as artifact-free ground truth $\{Y^*_v\}_{v=0}^9$. For each character, we reconstruct a 3D model from a single frontal view and project it to the same viewpoints to obtain artifact-prone projections $\{Z_v\}_{v=0}^9$. Each training sample consists of a poor input triplet $(Z_v, Z^{mask}_v, Z^{pos}_v)$ (RGB projection, foreground mask, and positional hint) paired with the corresponding GT rendering $Y^*_v$, yielding 14,980 poor-GT pairs for 2D projection refinement.
  }
  \label{fig:dataset}
\end{figure}

\subsection{SPECSIA-15K Dataset}
\label{dataset}
To learn priors for novel-view error correction, we require scalable supervision that connects
(i) artifact-prone projected results and (ii) corresponding artifact-free targets (Fig.~\ref{fig:dataset}).
Such scalable data is scarce for hand-drawn character animation, which has motivated character-specific, sample-wise optimization in prior work.
We fill this gap by introducing \textbf{SPECSIA-15K}, a paired dataset specialized for 2D projection refinement across viewpoints in drawing-based 3D animation.

\subsubsection{Data Source. }
We use drawing-style character assets from 3DBiCar \cite{luo2023rabit}.
For each character, we generate multi-view artifact-free targets by directly rendering the original asset, and multi-view artifact-prone projections by reconstructing a 3D character from a single reference view and re-projecting it to novel views.

\subsubsection{View Sampling and Pairing. }
For each character, we uniformly sample 10 viewpoints along yaw in $0^\circ \sim 360^\circ$ and render artifact-free targets $\{Y^*_v\}_{v=0}^9$.
We then select the frontal view $Y^*_0$ as the single-view reference input to an image-to-3D reconstruction pipeline (e.g., diffusion-based reconstruction and rigging~\cite{long2023wonder3d,10.1145/1276377.1276467}).
The reconstructed 3D model is projected into the same 10 viewpoints, yielding artifact-prone projections
$\{Z_v\}_{v=0}^9$.
In practice, $Z_v$ contains line/contour distortions, texture degradation, and view-dependent artifacts that are amplified in unseen viewpoints.
This paired setup aligns inputs and targets by viewpoint.
In addition to the RGB projection $Z_v$, we provide two auxiliary signals:
\begin{itemize}
    \item \textbf{Foreground mask} $Z^{mask}_v$: binary mask indicating valid object regions.
    \item \textbf{Positional hint} $Z^{pos}_v$: aligned map encoding canonical surface coordinates.
\end{itemize}
Each sample forms a triplet input $(Z_v, Z^{mask}_v, Z^{pos}_v)$ paired with target $Y^*_v$.

\subsubsection{Rendering Protocol and Auxiliary Signals. }
\paragraph{\textbf{Artifact-free targets $Y^*_v$. }}
We render 3DBiCar characters in Blender \cite{blender} using an orthographic camera and transparent background, producing RGBA PNG frames at $512 \times 512$ resolution.
To match diverse drawing-like appearances, we overlay a contour outline using Blender Freestyle as a style augmentation.
Specifically, we sample the outline thickness uniformly from $\{1,2,3,4\}$ pixels and sample the line color in grayscale from $[0,255]$ (black to white), then apply it within each rendered frame.
We use the same camera setup across views and uniformly rotate the camera rig around the character to obtain 10 yaw-spaced views.

\paragraph{\textbf{Artifact-prone projections $Z_v$. }}
To generate $Z_v$, we reconstruct a 3D character from the single reference view $Y^*_0$ using an image-to-3D model and apply rigging \cite{long2023wonder3d,10.1145/1276377.1276467}.
We then render the reconstructed character under the same set of viewpoints.
Because the reconstructed model can be imperfect, these projections naturally exhibit systematic projection errors such as missing textures, warped contours, and view-inconsistent patterns, particularly in novel views.

\paragraph{\textbf{Foreground mask $Z^{mask}_v$. }}
We generate $Z^{mask}_v$ by rendering a silhouette-only pass (emission material) for each view and extracting the alpha mask.
To reduce view-dependent misalignment between the projected mesh and the target silhouette, we optionally adjust camera/object alignment by searching small translations and pitch/roll offsets that maximize mask IoU with the target silhouette.
This alignment step improves the correspondence between $Z_v$ and $Y^*_v$ during pair construction without altering the underlying reconstruction pipeline.

\paragraph{\textbf{Positional hint $Z^{pos}_v$. }}
Inspired by the positional conditioning strategy used in DrawingSpinUp \cite{zhou2024drawingspinup} and Stylizing Video by Example \cite{10.1145/3306346.3323006}, we generate $Z^{pos}_v$ by encoding canonical mesh coordinates as colors.
Specifically, we min-max normalize mesh vertex coordinates and assign $(\tilde{x},\tilde{y},\tilde{z})\in[0,1]^3$ to vertex colors, then render it as an RGB map aligned with each projected frame.
This conditioning provides a spatial cue about where a patch originates on the character surface.

\subsubsection{Dataset Scale and Usage.}
SPECSIA-15K includes 1,498 characters and 10 viewpoints per character, resulting in 14,980 samples.
SPECSIA-15K provides poor-GT pairs tailored to learning robust priors for projection refinement.
By training on diverse characters and viewpoints, DraViE learns corrections that generalize beyond the reference view and improve stability in unseen viewpoints.

\begin{figure}
  \centering
  \includegraphics[width=1.0\textwidth]{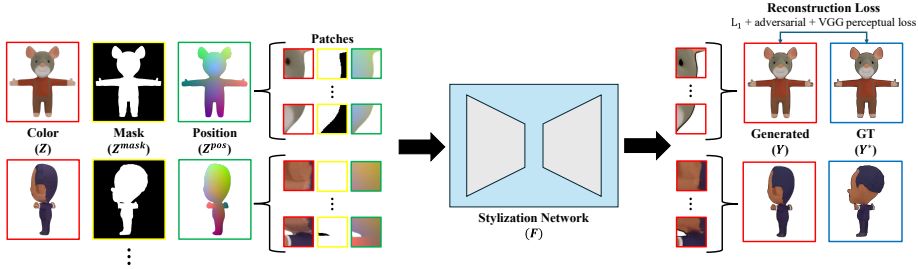}
  \caption{Pretraining DraViE model. Each character has 10 different views, and each RGB projection is paired with its mask and positional hint. These triplets are split into patches, and a stylization network is trained to reconstruct these patches. Before inference, the pre-trained DraViE model is optionally adapted to the input drawing.
  }
  \label{fig:pretrain}
\end{figure}

\subsection{Pre-training DraViE with Data-level Priors}
\label{pretrain}
We pre-train $F$ on SPECSIA-15K to learn general refinement priors for recurring artifact patterns in 3D renderings and 2D projections.
We follow the overall training scheme of DrawingSpinUp with minor adaptations to our setting \cite{zhou2024drawingspinup}.

\subsubsection{Training Objective. }
The model predicts $Y$ from $(Z, Z^{mask}, Z^{pos})$ and is supervised by $Y^*$, the corresponding artifact-free target rendering:
\begin{equation}
F(Z, Z^{mask}, Z^{pos}) \approx Y^* .
\end{equation}
Pre-training differs from sample-wise optimization in that it learns correction priors over a dataset distribution, improving generalization to unseen viewpoints.

\subsubsection{Patch-wise Training. }
Hand-drawn style is governed by local cues (e.g., stroke thickness and texture granularity), which are difficult to capture from full-resolution images. Moreover, training on full frames can bias the model toward dominant pose patterns (e.g., the canonical T-pose), leading to pose-specific overfitting; patch-wise training mitigates this issue by emphasizing local style cues. As Fig. \ref{fig:pretrain} shows, we sample a patch center guided by $Z^{mask}$ and crop patches of size $32 \times 32$ from ($Z$, $Z^{mask}$, $Z^{pos}$) and $Y^*$. The model predicts the refined patch $Y_{patch}$, and we supervise it with a combination of $L_1$ loss, adversarial loss, and VGG perceptual loss between $Y_{patch}$ and $Y^*_{patch}$. Each training step, we choose a patch from a random character and a random viewpoint.
At inference time, since $F$ is fully convolutional, it can process full-resolution inputs (not limited to $32 \times 32$), enabling practical refinement of full-frame projected renderings.

\subsection{Lightweight Adaptation (Optional)}
\label{adapt}
While the pre-trained model generalizes across diverse characters, test-time inputs may exhibit unique drawing styles (e.g., line thickness or color palette).
We optionally perform a lightweight adaptation to calibrate DraViE to the input character, improving character-specific appearance while preserving data-level correction priors.
Unlike full sample-wise optimization, this adaptation is computationally inexpensive and mitigates strong overfitting to the reference view.
Concretely, we initialize from the SPECSIA-15K-pretrained generator and fine-tune it on the input character for a single epoch using the same patch-wise objective as pre-training.
We keep the architecture and loss unchanged and only reduce the generator learning rate and the number of epochs.
We sample $32\times32$ patches from the character's input pose renderings (with the same auxiliary inputs) and optimize a combination of $L_1$ loss, VGG perceptual loss, and adversarial loss.
This brief update improves input-style alignment (colors/strokes/textures) for the given character while retaining the learned artifact-correction prior.

\subsection{Inference}
\label{inference}
Given an input drawing $X$ and motion sequence $\{P_i\}$, the 3D animation system produces projected frames $\{Z_i\}$ along with auxiliary signals $\{Z^{mask}_i, Z^{pos}_i\}$ for all animation timesteps in the sequence. DraViE refines each projected frame as:
\begin{equation}
Y_i = F\!\left(Z_i, Z^{mask}_i, Z^{pos}_i\right).
\end{equation}
By leveraging learned priors, DraViE reduces view-dependent artifacts, preserves drawing style, and improves temporal coherence across novel views.

\section{Experiments}
\subsection{Experimental Settings}
\subsubsection{Implementation Details. }
For image-to-3D reconstruction, we use Wonder3D \cite{long2023wonder3d}. For rigging, we use Adobe Mixamo\footnote{We use the online platform (\url{https://www.mixamo.com}, accessed: 2026-03-09).} with a 65-joint skeleton.
Unless otherwise specified, we render sequences with the same camera and rendering settings across methods to ensure a fair and controlled comparison setting.

\subsubsection{Data and Baselines. }
SPECSIA-15K contains multi-view projections of 1,498 3DBiCar characters with 10 views per character. We use a character-disjoint split of 1,298/100/100 characters for train/validation/test, respectively, so that no character identity is shared between pre-training and evaluation.
For evaluation, we render a near-frontal view for each test character as the single reference input. This choice reflects practical user inputs that are close to frontal but not strictly frontal.
To further assess generalization to hand-drawn inputs, we additionally select 20 textured characters from Amateur Drawings \cite{10.1145/3592788}, excluding characters with nearly monochromatic body textures.
Each test character is driven by 20 motion sequences (10 frontal-dominant and 10 novel-view motions\footnote{Detailed Mixamo motions are listed in Appendix~A.}), resulting in $(100+20)\times 20=2,400$ animation samples in total.
We compare against two drawing-based 3D animation pipelines, DrawingSpinUp (DSU) \cite{zhou2024drawingspinup} and the Occlusion-robust Stylization Framework (OSF) \cite{yoon2025occlusion}.
For all methods, we use identical 3D reconstruction and rigging inputs to isolate the effect of the 2D post-correction module. Additional clean-target, dataset-split, view-sector, efficiency, protocol, and qualitative analyses are provided in the supplementary material.

\subsection{Evaluation Metrics}
We evaluate the output video quality in terms of temporal coherence, fidelity, and overall realism.
For temporal coherence, we compute (i) the average cosine similarity of adjacent-frame CLIP image embeddings \cite{pmlr-v139-radford21a}, capturing semantic/appearance consistency over time, and (ii) the average SSIM \cite{1284395} between adjacent frames, measuring structural stability.
For fidelity, we report LPIPS \cite{8578166} between each generated frame and the reference appearance image (the character texture), assessing perceptual identity preservation.
For overall realism, we compute FID \cite{NIPS2017_8a1d6947} between the set of reference images and the set of generated frames (sampled from all videos), measuring distribution-level visual quality.
All metrics are averaged over 10 runs with different random seeds.

\subsection{Experimental Results}

\begin{figure}[t]
  \centering
  \includegraphics[width=1.0\textwidth]{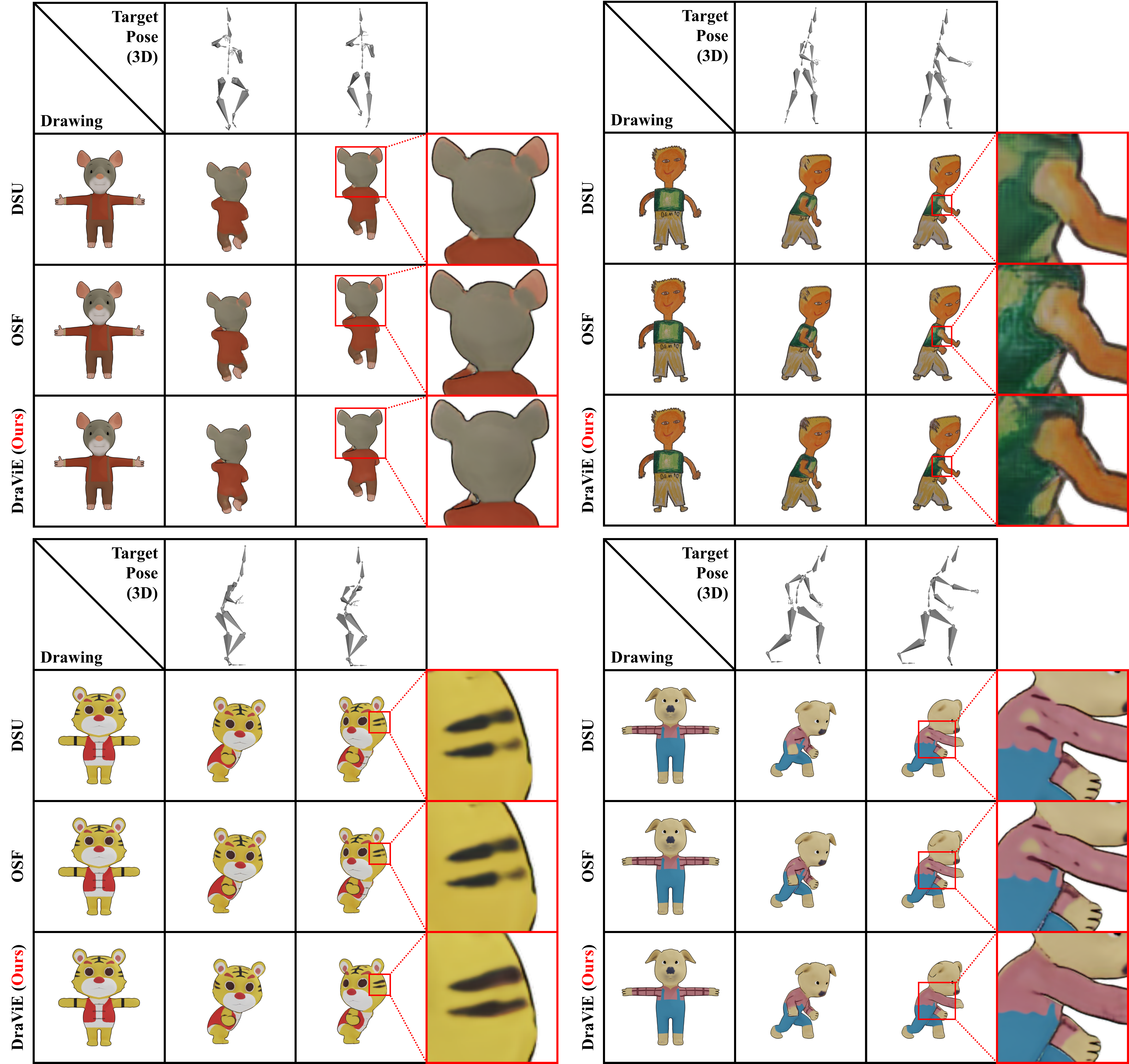}
  \caption{Qualitative comparisons across different drawing-based 3D animation models. The red boxes highlight zoomed views of the stylization applied to animated drawing objects. DSU: DrawingSpinUp \cite{zhou2024drawingspinup}, OSF: Occlusion-robust Stylization Framework \cite{yoon2025occlusion}.
  }
  \label{fig:Qualitative}
\end{figure}

\subsubsection{Qualitative Comparisons. }
Fig. \ref{fig:Qualitative} compares outputs across different 3D animation systems. We evaluate the methods under dynamic motions (e.g., turning and punching) on diverse characters.
All methods preserve the input drawing style. However, the artifact patterns differ.
DrawingSpinUp (DSU) \cite{zhou2024drawingspinup} tends to lose contour details under large motions, resulting in blurred textures.
Both DSU and the Occlusion-robust Stylization Framework (OSF) \cite{yoon2025occlusion} show speckle-like artifacts amplified in novel views due to rendering and projection errors.
In contrast, DraViE reduces both blurring and speckling, effectively correcting projection-induced artifacts that prior methods often fail to address. \footnote{More qualitative results are provided in Appendix~B.}

\begin{table}[t]
    \caption{Quantitative evaluations on drawing-based 3D animation models. DSU: DrawingSpinUp \cite{zhou2024drawingspinup}, OSF:  Occlusion-robust Stylization Framework \cite{yoon2025occlusion}.}
    \label{tab:quantitative}
    \centering
    \setlength{\tabcolsep}{5pt}
    \renewcommand{\arraystretch}{1.10}

    \resizebox{\linewidth}{!}{%
        \begin{tabular}{lcccccccc}
            \toprule
            & \multicolumn{4}{c}{\textbf{Frontal-view}} & \multicolumn{4}{c}{\textbf{Novel-view}}\\
            \cmidrule(lr){2-5}\cmidrule(lr){6-9}
            \textbf{Method}
            & \textbf{CLIP}$\uparrow$
            & \textbf{SSIM}$\uparrow$
            & \textbf{LPIPS}$\downarrow$
            & \textbf{FID}$\downarrow$
            & \textbf{CLIP}$\uparrow$
            & \textbf{SSIM}$\uparrow$
            & \textbf{LPIPS}$\downarrow$
            & \textbf{FID}$\downarrow$ \\
            
            \midrule
            DSU \cite{zhou2024drawingspinup} & 0.902 & 0.840 & 0.212 &  205.19 & 0.847 & 0.833 & 0.250 & 280.83\\
            OSF \cite{yoon2025occlusion} & 0.905 & 0.841 & 0.211 & 202.24 & 0.850 & 0.835 & 0.249 & 270.21\\
            \rowcolor{green!10} \textbf{DraViE (Ours)} & \textbf{0.905} & \textbf{0.846} & \textbf{0.208} & \textbf{201.31} & \textbf{0.859} & \textbf{0.840} & \textbf{0.245} & \textbf{206.13} \\
            \bottomrule
        \end{tabular}%
    }
\end{table}

\subsubsection{Quantitative Results.}
Table \ref{tab:quantitative} reports quantitative comparisons on drawing-based 3D animation systems. We evaluated the animation quality on motions dominated by frontal views and novel views.
DraViE achieves performance comparable to prior methods on frontal-view animations, indicating that our correction does not degrade quality in the observed view.
More importantly, DraViE yields consistent gains in novel views across all metrics, with higher CLIP \cite{pmlr-v139-radford21a} and SSIM \cite{1284395} and lower LPIPS \cite{8578166} and FID \cite{NIPS2017_8a1d6947}.
The large reduction in novel-view FID suggests that DraViE produces frames that are closer to the reference distribution and contain fewer projection-induced artifacts.
These quantitative improvements align with the qualitative comparisons in Fig. \ref{fig:Qualitative}.

\begin{table}[t]
    \caption{Cross-pipeline generalization and blinded user preference.
    Left: each triplet reports DSU/OSF/DraViE using the same SPECSIA-pretrained checkpoint and the same adaptation protocol, without backend-specific retraining.
    Right: preference values are percentages from a 30-person blinded A/B study.
    }
    \label{tab:cross_user}
    \centering
    \footnotesize
    \setlength{\tabcolsep}{2.5pt}
    \renewcommand{\arraystretch}{1.05}
    \begin{minipage}{0.61\linewidth}
    \centering
    \resizebox{\linewidth}{!}{%
    \begin{tabular}{lccc}
        \toprule
        \textbf{Upstream} & \textbf{CLIP}$\uparrow$ & \textbf{LPIPS}$\downarrow$ & \textbf{FID}$\downarrow$\\
        \midrule
        Wonder3D & .874/.877/\textbf{.882} & .231/.230/\textbf{.225} & 243/236/\textbf{203}\\
        InstantMesh & .831/.835/\textbf{.840} & .277/.275/\textbf{.272} & 242/233/\textbf{205}\\
        CRM & .807/.810/\textbf{.848} & .272/.271/\textbf{.267} & 271/266/\textbf{239}\\
        \bottomrule
    \end{tabular}}
    \end{minipage}
    \hfill
    \begin{minipage}{0.35\linewidth}
    \centering
    \resizebox{\linewidth}{!}{%
    \begin{tabular}{lccc}
        \toprule
        \textbf{Method} & \textbf{Art.} & \textbf{Style} & \textbf{Overall}\\
        \midrule
        Prior & 26.7 & 40.0 & 13.3\\
        \textbf{DraViE} & \textbf{73.3} & \textbf{60.0} & \textbf{86.7}\\
        \bottomrule
    \end{tabular}}
    \end{minipage}
\end{table}

\subsubsection{Cross-pipeline Generalization and User Preference.}
To examine whether DraViE learns a transferable correction prior rather than a backend-specific denoiser, we apply the same SPECSIA-pretrained checkpoint to projections generated by Wonder3D~\cite{long2023wonder3d}, InstantMesh~\cite{xu2024instantmesh}, and CRM~\cite{wang2024crm}, using the same lightweight adaptation protocol without backend-specific retraining.
As shown in Table~\ref{tab:cross_user}, DraViE consistently improves over DSU and OSF across all upstream backends, suggesting that the learned prior transfers across different reconstruction artifacts.
The upstream models produce different artifact patterns; some projections contain texture bleeding near front/back surface boundaries, while others show clay-like or over-smoothed surface appearance.
Despite these differences, DraViE improves the output by targeting projection-level errors such as speckling, boundary corruption, and texture inconsistency, rather than relying on a specific reconstruction backend.
We also conduct a 30-person blinded A/B preference study with hidden method names and randomized order.
Each participant compares DraViE against a randomly selected prior method under artifact reduction, style preservation, and overall visual quality criteria.
Participants prefer DraViE across all three criteria, which supports the quantitative gains and indicates that the improvements are perceptually visible.
Cross-pipeline qualitative examples, descriptor-space backend discrepancy, view-sector analysis, efficiency results, and additional analyses are provided in the supplementary material.

\begin{figure}
  \centering
  \includegraphics[width=1.0\textwidth]{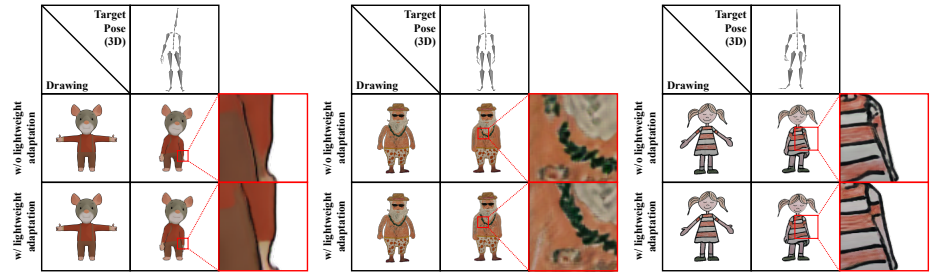}
  \caption{Qualitative comparison of DraViE with and without the lightweight adaptation. Without adaptation, the pre-trained prior removes projection artifacts but often deviates from the input drawing appearance. It leads to mismatched colors and over-smoothed or altered strokes (e.g., hand color, leaf pattern, and stripe line thickness).
  }
  \label{fig:ablation_lightweight}
\end{figure}

\subsection{Ablation Study}

\begin{table}[t]
    \caption{Quantitative ablations on (i) lightweight adaptation and (ii) positional hint.}
    \label{tab:quantitative_ablation}
    \centering
    \setlength{\tabcolsep}{5pt}
    \renewcommand{\arraystretch}{1.10}

    \resizebox{\linewidth}{!}{%
        \begin{tabular}{lcccccccc}
            \toprule
            & \multicolumn{4}{c}{\textbf{Frontal-view}} & \multicolumn{4}{c}{\textbf{Novel-view}}\\
            \cmidrule(lr){2-5}\cmidrule(lr){6-9}
            \textbf{Setting}
            & \textbf{CLIP}$\uparrow$
            & \textbf{SSIM}$\uparrow$
            & \textbf{LPIPS}$\downarrow$
            & \textbf{FID}$\downarrow$
            & \textbf{CLIP}$\uparrow$
            & \textbf{SSIM}$\uparrow$
            & \textbf{LPIPS}$\downarrow$
            & \textbf{FID}$\downarrow$ \\
            \midrule
            w/o LA & 0.900 & 0.840 & 0.212 & 207.15 & 0.838 & 0.834 & 0.250 & 221.48\\
            \textbf{w/ LA} & \textbf{0.905} & \textbf{0.846} & \textbf{0.208} & \textbf{201.31} & \textbf{0.859} & \textbf{0.840} & \textbf{0.245} & \textbf{206.13} \\
            \midrule
            w/o $Z^{pos}$ & 0.900 & 0.845 & 0.212 & 205.81 & 0.844 & 0.839 & 0.250 & 318.39\\
            \textbf{w/ $Z^{pos}$} & \textbf{0.905} & \textbf{0.846} & \textbf{0.208} & \textbf{201.31} & \textbf{0.859} & \textbf{0.840} & \textbf{0.245} & \textbf{206.13} \\
            \bottomrule
        \end{tabular}%
    }
\end{table}

\subsubsection{Lightweight Adaptation. }
We introduce lightweight adaptation to better align DraViE with the appearance of the input drawing (e.g., stroke thickness, local textures, and color tones) while retaining the pre-trained correction prior.
Table \ref{tab:quantitative_ablation} shows that even without adaptation, DraViE already provides strong correction performance, especially on the frontal view. It indicates that the pre-trained model already captures common projection error patterns and can suppress view-dependent artifacts.
However, the outputs can deviate from the input drawing appearance in novel views, which is consistent with lower CLIP \cite{pmlr-v139-radford21a} and higher LPIPS \cite{8578166} and FID \cite{NIPS2017_8a1d6947}.
With lightweight adaptation, DraViE improves perceptual alignment to the input, and the gains are more pronounced in novel views (e.g., CLIP: 0.838$\rightarrow$0.859, FID: 221.48$\rightarrow$206.13).
Fig. \ref{fig:ablation_lightweight} qualitatively supports this trend. Without adaptation, the model tends to over-smooth textures and strokes, whereas adaptation better preserves detailed textures (e.g., hand colors, leaf patterns, and stripe strokes) while still correcting projection artifacts.

\begin{figure}
  \centering
  \includegraphics[width=1.0\textwidth]{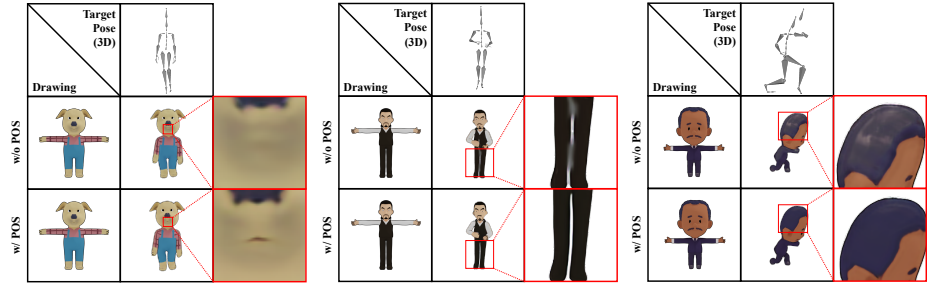}
  \caption{Qualitative comparison of DraViE trained with and without the positional hint. Without positional hint, the model often fails to recover fine structures (e.g., mouth) and leaves blurring artifacts at specific regions (e.g., hair, legs).}
  \label{fig:ablation_pos}
\end{figure}

\subsubsection{Input Signal Ablation.}
DraViE uses two auxiliary inputs, mask and positional hint. The mask is essential for patch-wise training, while the positional hint provides the spatial location of each input patch. In this ablation, we show the effect of the positional hint by training DraViE with and without it.
Fig. \ref{fig:ablation_pos} shows that removing the positional hint significantly degrades output quality. Without positional information, the model observes only local appearances and cannot distinguish
where a patch comes from on the character. As a result, the model tends to apply conservative smoothing over corrupted regions and fails to reconstruct geometric structures (e.g., mouth) and often leaves speckling artifacts induced from the projections (e.g., around hair and legs).
In contrast, conditioning on the positional hint enables DraViE to learn location-aware correction priors, leading to sharper structure recovery and cleaner textures under viewpoint changes. Table \ref{tab:quantitative_ablation} also shows that removing the positional hint degrades the quality of animation, confirming its role in stable refinement.

\section{Limitations}

\begin{wrapfigure}{r}{0.44\textwidth}
  \centering
  \includegraphics[width=0.40\textwidth]{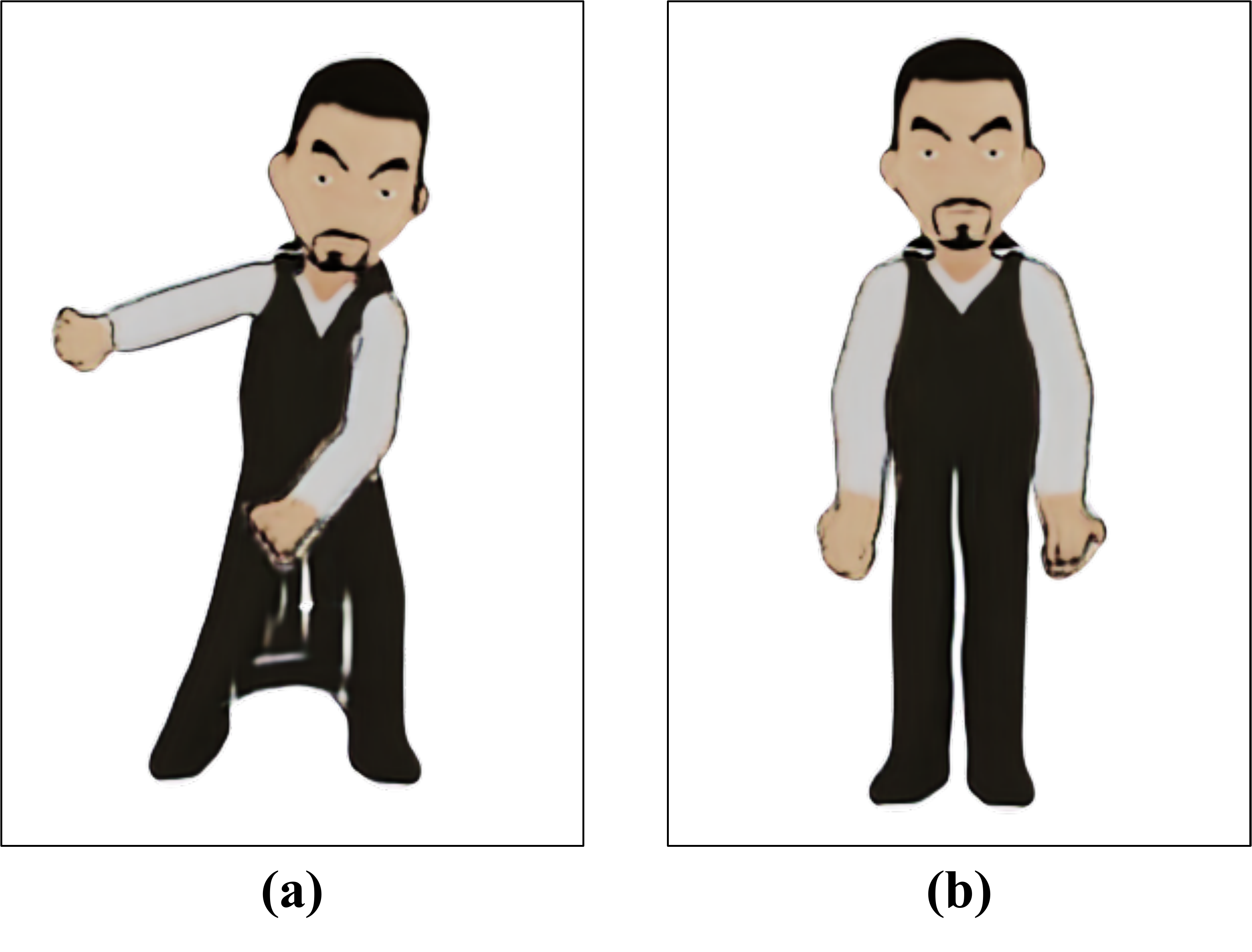}
  \caption{Failure case. (a) Fused legs. (b) Hallucinated gap.}
  \label{fig:limit}
\end{wrapfigure}

DraViE is designed as a modular 2D post-correction component rather than a replacement for upstream 3D reconstruction, rigging, or temporal modeling. This makes the method easy to attach to existing drawing-based animation pipelines, but it also bounds the correction quality by the rendered proxy and limits the algorithmic scope of the model. In particular, DraViE cannot fully recover from severe upstream failures such as incorrect topology, missing large regions, or erroneous rigging.

For example, when the reconstructed mesh merges the legs into a single region, DraViE cannot reliably reconstruct separation. In some cases, the learned prior may hallucinate a leg gap and refine the projection to look separated, despite the input exhibiting a fused-leg structure (Fig.~\ref{fig:limit}). This failure highlights a fundamental limitation of 2D post-correction: it may favor a plausible silhouette prior over the true input geometry when upstream topology is incorrect.

SPECSIA-15K is constructed with an offline rendering, reconstruction, and rigging pipeline. Although the supplementary material reports cross-pipeline evaluations, clean-target, and dataset-split, the dataset may encode systematic biases from the source assets and upstream reconstruction backend. Finally, DraViE processes frames independently and does not use explicit temporal attention or optical flow. The shared 3D proxy and positional hints improve stability in practice, but rapid rotations, large occlusions, or high-frequency details that are missing in the projection may still lead to flickering or over-smoothed details.

\section{Conclusion}
We introduced \textbf{SPECSIA-15K}, a paired multi-view stylization dataset for projection refinement in drawing-based 3D animation, and \textbf{DraViE}, a lightweight plug-and-play refinement module trained with data-level correction priors.

SPECSIA-15K provides scalable supervision for correcting projection artifacts under dynamic motion and varying viewpoints, moving beyond sample-wise refinement that can overfit to the observed view.
By explicitly pairing artifact-prone projections with clean multi-view targets, SPECSIA-15K enables learning correction priors that are difficult to obtain from single-character optimization alone.
Building on this dataset, DraViE improves novel-view fidelity and temporal coherence while preserving the input drawing style, and integrates into existing animation pipelines without modifying their core components.

These results suggest that dataset-level supervision is a practical direction for improving robustness in drawing-based 3D animation systems.
At the same time, projection-level refinement cannot fully repair severe reconstruction or rigging failures, motivating future work on less biased data construction, stronger 3D reconstruction, and explicit temporal modeling.

We will release the training/evaluation code, split lists, rendering parameters, metadata, and redistributable SPECSIA noisy/clean pairs when permitted by the corresponding third-party licenses.

\section*{Acknowledgements}
This work was supported by Institute for Information \& communications Technology Planning \& Evaluation (IITP) grant funded by the Korea government(MSIT) (No.RS-2021-II211381, Development of Causal AI through Video Understanding and Reinforcement Learning, and Its Applications to Real Environments) and (No.RS-2022-II220184, Development and Study of AI Technologies to Inexpensively Conform to Evolving Policy on Ethics)

\bibliographystyle{splncs04}
\bibliography{main}

@String(CVPR  = {IEEE Conf. Comput. Vis. Pattern Recog.})

@String(ICCV  = {Int. Conf. Comput. Vis.})

@String(ECCV  = {Eur. Conf. Comput. Vis.})

@String(NeurIPS = {Adv. Neural Inform. Process. Syst.})

@String(ICLR  = {Int. Conf. Learn. Represent.})

@String(ACCV  = {Asian Conf. Comput. Vis.})

@String(TOG   = {ACM Trans. Graph.})

@inproceedings{Siarohin_2019_NeurIPS,
  author    = {Siarohin, Aliaksandr and Lathuili\`ere, St\'ephane and Tulyakov, Sergey and Ricci, Elisa and Sebe, Nicu},
  title     = {First Order Motion Model for Image Animation},
  booktitle = {Advances in Neural Information Processing Systems},
  volume    = {32},
  year      = {2019}
}

@inproceedings{hu2023animateanyone,
  title={Animate Anyone: Consistent and Controllable Image-to-Video Synthesis for Character Animation},
  author={Hu, Li and Gao, Xin and Zhang, Peng and Sun, Ke and Zhang, Bang and Bo, Liefeng},
  booktitle={Proceedings of the IEEE/CVF Conference on Computer Vision and Pattern Recognition (CVPR)},
  pages={8153--8163},
  year={2024}
}

@inproceedings{zhou2024drawingspinup,
  author    = {Zhou, Jie and Xiao, Chufeng and Lam, Miu-Ling and Fu, Hongbo},
  title     = {DrawingSpinUp: 3D Animation from Single Character Drawings},
  booktitle = {SIGGRAPH Asia 2024 Conference Papers},
  publisher = {Association for Computing Machinery},
  address   = {New York, NY, USA},
  year      = {2024}
}

@inproceedings{yoon2025occlusion,
  title={Occlusion-robust Stylization for Drawing-based 3D Animation},
  author={Yoon, Sunjae and Koo, Gwanhyeong and Lee, Younghwan and Hong, Ji Woo and Yoo, Chang D.},
  booktitle={Proceedings of the IEEE/CVF International Conference on Computer Vision (ICCV)},
  pages={12263--12273},
  year={2025}
}

@inproceedings{xu2023magicanimate,
  title={MagicAnimate: Temporally Consistent Human Image Animation using Diffusion Model},
  author={Xu, Zhongcong and Zhang, Jianfeng and Liew, Jun Hao and Yan, Hanshu and Liu, Jia-Wei and Zhang, Chenxu and Feng, Jiashi and Shou, Mike Zheng},
  booktitle={Proceedings of the IEEE/CVF Conference on Computer Vision and Pattern Recognition (CVPR)},
  pages={1481--1490},
  year={2024}
}

@inproceedings{guo2023animatediff,
  title     = {AnimateDiff: Animate Your Personalized Text-to-Image Diffusion Models without Specific Tuning},
  author    = {Guo, Yuwei and Yang, Ceyuan and Rao, Anyi and Liang, Zhengyang and Wang, Yaohui and Qiao, Yu and Agrawala, Maneesh and Lin, Dahua and Dai, Bo},
  booktitle = {The Twelfth International Conference on Learning Representations},
  year      = {2024}
}

@inproceedings{tang2023dreamgaussian,
  title     = {DreamGaussian: Generative Gaussian Splatting for Efficient 3D Content Creation},
  author    = {Tang, Jiaxiang and Ren, Jiawei and Zhou, Hang and Liu, Ziwei and Zeng, Gang},
  booktitle = {The Twelfth International Conference on Learning Representations},
  year      = {2024}
}

@inproceedings{long2023wonder3d,
  title={Wonder3D: Single Image to 3D using Cross-Domain Diffusion},
  author={Long, Xiaoxiao and Guo, Yuan-Chen and Lin, Cheng and Liu, Yuan and Dou, Zhiyang and Liu, Lingjie and Ma, Yuexin and Zhang, Song-Hai and Habermann, Marc and Theobalt, Christian and Wang, Wenping},
  booktitle={Proceedings of the IEEE/CVF Conference on Computer Vision and Pattern Recognition (CVPR)},
  pages={9970--9980},
  year={2024}
}

@inproceedings{luo2023rabit,
  title     = {RaBit: Parametric Modeling of 3D Biped Cartoon Characters with a Topological-consistent Dataset},
  author    = {Luo, Zhongjin and Cai, Shengcai and Dong, Jinguo and Ming, Ruibo and Qiu, Liangdong and Zhan, Xiaohang and Han, Xiaoguang},
  booktitle = {Proceedings of the IEEE/CVF Conference on Computer Vision and Pattern Recognition (CVPR)},
  pages     = {12825--12835},
  year      = {2023}
}

@article{aberman2019learning,
  author = {Aberman, Kfir and Wu, Rundi and Lischinski, Dani and Chen, Baoquan and Cohen-Or, Daniel},
  title = {Learning Character-Agnostic Motion for Motion Retargeting in 2D},
  journal = {ACM Transactions on Graphics (TOG)},
  volume = {38},
  number = {4},
  pages = {75},
  year = {2019},
  publisher = {ACM}
}

@article{albahar2021pose,
  title     = {Pose with {S}tyle: {D}etail-Preserving Pose-Guided Image Synthesis with Conditional StyleGAN},
  author    = {AlBahar, Badour and Lu, Jingwan and Yang, Jimei and Shu, Zhixin and Shechtman, Eli and Huang, Jia-Bin},
  journal   = {ACM Transactions on Graphics},
  volume    = {40},
  number    = {6},
  articleno = {218},
  numpages  = {12},
  year      = {2021},
  publisher = {Association for Computing Machinery}
}

@inproceedings{10.1145/1057432.1057456,
    author = {Sorkine, O. and Cohen-Or, D. and Lipman, Y. and Alexa, M. and R\"{o}ssl, C. and Seidel, H.-P.},
    title = {Laplacian surface editing},
    year = {2004},
    isbn = {3905673134},
    publisher = {Association for Computing Machinery},
    address = {New York, NY, USA},
    booktitle = {Proceedings of the 2004 Eurographics/ACM SIGGRAPH Symposium on Geometry Processing},
    pages = {175--184},
    numpages = {10},
    location = {Nice, France},
    series = {SGP '04}
}

@inproceedings{koo2025flowdrag,
  title={FlowDrag: 3D-aware Drag-based Image Editing with Mesh-guided Deformation Vector Flow Fields},
  author={Koo, Gwanhyeong and Yoon, Sunjae and Lee, Younghwan and Hong, Ji Woo and Yoo, Chang D.},
  booktitle={Proceedings of the 42nd International Conference on Machine Learning},
  series={Proceedings of Machine Learning Research},
  volume={267},
  pages={31467--31485},
  publisher={PMLR},
  year={2025}
}

@inproceedings{yoon2024tpc,
  title={TPC: Test-time Procrustes Calibration for Diffusion-based Human Image Animation},
  author={Yoon, Sunjae and Koo, Gwanhyeong and Lee, Younghwan and Yoo, Chang D.},
  booktitle={Advances in Neural Information Processing Systems (NeurIPS)},
  volume={37},
  pages={118654--118677},
  year={2024}
}

@inproceedings{liu2023syncdreamer,
  title     = {SyncDreamer: Generating Multiview-consistent Images from a Single-view Image},
  author    = {Liu, Yuan and Lin, Cheng and Zeng, Zijiao and Long, Xiaoxiao and Liu, Lingjie and Komura, Taku and Wang, Wenping},
  booktitle = {The Twelfth International Conference on Learning Representations},
  year      = {2024}
}

@inproceedings{gatys2015neuralalgorithmartisticstyle,
  title={Image Style Transfer Using Convolutional Neural Networks},
  author={Gatys, Leon A. and Ecker, Alexander S. and Bethge, Matthias},
  booktitle={Proceedings of the IEEE Conference on Computer Vision and Pattern Recognition (CVPR)},
  pages={2414--2423},
  year={2016}
}

@inproceedings{isola2017image,
  title     = {Image-to-Image Translation with Conditional Adversarial Networks},
  author    = {Isola, Phillip and Zhu, Jun-Yan and Zhou, Tinghui and Efros, Alexei A.},
  booktitle = {Proceedings of the IEEE Conference on Computer Vision and Pattern Recognition (CVPR)},
  pages     = {1125--1134},
  year      = {2017}
}

@inproceedings{CycleGAN2017,
  title     = {Unpaired Image-to-Image Translation using Cycle-Consistent Adversarial Networks},
  author    = {Zhu, Jun-Yan and Park, Taesung and Isola, Phillip and Efros, Alexei A.},
  booktitle = {Proceedings of the IEEE International Conference on Computer Vision (ICCV)},
  pages     = {2223--2232},
  year      = {2017}
}

@inproceedings{10.5555/3495724.3496298,
  author    = {Ho, Jonathan and Jain, Ajay and Abbeel, Pieter},
  title     = {Denoising Diffusion Probabilistic Models},
  booktitle = {Advances in Neural Information Processing Systems},
  volume    = {33},
  pages     = {6840--6851},
  year      = {2020}
}

@ARTICLE{9887996,
  author={Saharia, Chitwan and Ho, Jonathan and Chan, William and Salimans, Tim and Fleet, David J. and Norouzi, Mohammad},
  journal={IEEE Transactions on Pattern Analysis and Machine Intelligence}, 
  title={Image Super-Resolution via Iterative Refinement}, 
  year={2023},
  volume={45},
  number={4},
  pages={4713-4726},
}

@InProceedings{Choi_2021_ICCV,
    author    = {Choi, Jooyoung and Kim, Sungwon and Jeong, Yonghyun and Gwon, Youngjune and Yoon, Sungroh},
    title     = {ILVR: Conditioning Method for Denoising Diffusion Probabilistic Models},
    booktitle = {Proceedings of the IEEE/CVF International Conference on Computer Vision (ICCV)},
    month     = {October},
    year      = {2021},
    pages     = {14367-14376}
}

@inbook{Ruder_2016,
   title={Artistic Style Transfer for Videos},
   ISBN={9783319458861},
   ISSN={1611-3349},
   booktitle={Pattern Recognition},
   publisher={Springer International Publishing},
   author={Ruder, Manuel and Dosovitskiy, Alexey and Brox, Thomas},
   year={2016},
   pages={26--36}
}

@inproceedings{conf/accv/GaoGZY18,
  author = {Gao, Chang and Gu, Derun and Zhang, Fangjun and Yu, Yizhou},
  booktitle = {ACCV (6)},
  editor = {Jawahar, C. V. and Li, Hongdong and Mori, Greg and Schindler, Konrad},
  isbn = {978-3-030-20876-9},
  pages = {637-653},
  publisher = {Springer},
  series = {Lecture Notes in Computer Science},
  title = {ReCoNet: Real-Time Coherent Video Style Transfer Network.},
  volume = 11366,
  year = 2018
}

@inproceedings{conf/iclr/SongME21,
  author       = {Jiaming Song and
                  Chenlin Meng and
                  Stefano Ermon},
  title        = {Denoising Diffusion Implicit Models},
  booktitle    = {9th International Conference on Learning Representations, {ICLR} 2021,
                  Virtual Event, Austria, May 3-7, 2021},
  publisher    = {OpenReview.net},
  year         = {2021},
}

@InProceedings{pmlr-v139-nichol21a,
  title = 	 {Improved Denoising Diffusion Probabilistic Models},
  author =       {Nichol, Alexander Quinn and Dhariwal, Prafulla},
  booktitle = 	 {Proceedings of the 38th International Conference on Machine Learning},
  pages = 	 {8162--8171},
  year = 	 {2021},
  editor = 	 {Meila, Marina and Zhang, Tong},
  volume = 	 {139},
  series = 	 {Proceedings of Machine Learning Research},
  month = 	 {18--24 Jul},
  publisher =    {PMLR},
}

@article{10.1145/3592788,
author = {Smith, Harrison Jesse and Zheng, Qingyuan and Li, Yifei and Jain, Somya and Hodgins, Jessica K.},
title = {A Method for Animating Children’s Drawings of the Human Figure},
year = {2023},
issue_date = {June 2023},
publisher = {Association for Computing Machinery},
address = {New York, NY, USA},
volume = {42},
number = {3},
issn = {0730-0301},
journal = {ACM Trans. Graph.},
month = {jun},
articleno = {32},
numpages = {15}
}

@InProceedings{pmlr-v139-radford21a,
  title = 	 {Learning Transferable Visual Models From Natural Language Supervision},
  author =       {Radford, Alec and Kim, Jong Wook and Hallacy, Chris and Ramesh, Aditya and Goh, Gabriel and Agarwal, Sandhini and Sastry, Girish and Askell, Amanda and Mishkin, Pamela and Clark, Jack and Krueger, Gretchen and Sutskever, Ilya},
  booktitle = 	 {Proceedings of the 38th International Conference on Machine Learning},
  pages = 	 {8748--8763},
  year = 	 {2021},
  editor = 	 {Meila, Marina and Zhang, Tong},
  volume = 	 {139},
  series = 	 {Proceedings of Machine Learning Research},
  month = 	 {18--24 Jul},
  publisher =    {PMLR},
}

@ARTICLE{1284395,
  author={Zhou Wang and Bovik, A.C. and Sheikh, H.R. and Simoncelli, E.P.},
  journal={IEEE Transactions on Image Processing}, 
  title={Image quality assessment: from error visibility to structural similarity}, 
  year={2004},
  volume={13},
  number={4},
  pages={600-612},
}

@inproceedings{NIPS2017_8a1d6947,
  author    = {Heusel, Martin and Ramsauer, Hubert and Unterthiner, Thomas and Nessler, Bernhard and Hochreiter, Sepp},
  title     = {{GANs} Trained by a Two Time-Scale Update Rule Converge to a Local Nash Equilibrium},
  booktitle = {Advances in Neural Information Processing Systems},
  volume    = {30},
  pages     = {6626--6637},
  year      = {2017}
}

@INPROCEEDINGS{8578166,
  author={Zhang, Richard and Isola, Phillip and Efros, Alexei A. and Shechtman, Eli and Wang, Oliver},
  booktitle={2018 IEEE/CVF Conference on Computer Vision and Pattern Recognition}, 
  title={The Unreasonable Effectiveness of Deep Features as a Perceptual Metric}, 
  year={2018},
  volume={},
  number={},
  pages={586-595},
}

@article{10.1145/1276377.1276467,
author = {Baran, Ilya and Popovi\'{c}, Jovan},
title = {Automatic rigging and animation of 3D characters},
year = {2007},
issue_date = {July 2007},
publisher = {Association for Computing Machinery},
address = {New York, NY, USA},
volume = {26},
number = {3},
issn = {0730-0301},
journal = {ACM Trans. Graph.},
month = jul,
pages = {72--es},
numpages = {8}
}

@InProceedings{10.1007/978-3-319-24574-4_28,
author="Ronneberger, Olaf
and Fischer, Philipp
and Brox, Thomas",
editor="Navab, Nassir
and Hornegger, Joachim
and Wells, William M.
and Frangi, Alejandro F.",
title="U-Net: Convolutional Networks for Biomedical Image Segmentation",
booktitle="Medical Image Computing and Computer-Assisted Intervention -- MICCAI 2015",
year="2015",
publisher="Springer International Publishing",
address="Cham",
pages="234--241",
isbn="978-3-319-24574-4"
}

@article{10.1145/3306346.3323006,
author = {Jamri\v{s}ka, Ond\v{r}ej and Sochorov\'{a}, \v{S}\'{a}rka and Texler, Ond\v{r}ej and Luk\'{a}\v{c}, Michal and Fi\v{s}er, Jakub and Lu, Jingwan and Shechtman, Eli and S\'{y}kora, Daniel},
title = {Stylizing video by example},
year = {2019},
issue_date = {August 2019},
publisher = {Association for Computing Machinery},
address = {New York, NY, USA},
volume = {38},
number = {4},
issn = {0730-0301},
journal = {ACM Trans. Graph.},
month = jul,
articleno = {107},
numpages = {11}
}

@misc{blender,
  author       = {{Blender Foundation}},
  title        = {Blender},
  howpublished = {\url{https://www.blender.org/}},
  year         = {2024},
  note         = {Version 3.6.14, accessed: 2026-03-09}
}

@article{mo2024ric,
  title={RIC-CNN: Rotation-invariant coordinate convolutional neural network},
  author={Mo, Hanlin and Zhao, Guoying},
  journal={Pattern Recognition},
  volume={146},
  pages={109994},
  year={2024},
  publisher={Elsevier}
}

@inproceedings{simonyan2015deepconvolutionalnetworkslargescale,
  title={Very Deep Convolutional Networks for Large-Scale Image Recognition},
  author={Simonyan, Karen and Zisserman, Andrew},
  booktitle={International Conference on Learning Representations (ICLR)},
  year={2015}
}

@inproceedings{kingma2015adam,
  author    = {Kingma, Diederik P. and Ba, Jimmy},
  title     = {Adam: A Method for Stochastic Optimization},
  booktitle = {International Conference on Learning Representations (ICLR)},
  year      = {2015}
}

@inproceedings{cherti2023reproducible,
  title={Reproducible scaling laws for contrastive language-image learning},
  author={Cherti, Mehdi and Beaumont, Romain and Wightman, Ross and Wortsman, Mitchell and Ilharco, Gabriel and Gordon, Cade and Schuhmann, Christoph and Schmidt, Ludwig and Jitsev, Jenia},
  booktitle={Proceedings of the IEEE/CVF conference on computer vision and pattern recognition},
  pages={2818--2829},
  year={2023}
}

@misc{Mixamo,
  author       = {{Adobe}},
  title        = {Mixamo},
  howpublished = {\url{https://www.mixamo.com}},
  note         = {accessed: 2026-03-09}
}

@inproceedings{koo2024flexiedit,
  title     = {FlexiEdit: Frequency-Aware Latent Refinement for Enhanced Non-Rigid Editing},
  author    = {Koo, Gwanhyeong and Yoon, Sunjae and Hong, Ji Woo and Yoo, Chang D.},
  booktitle = {Computer Vision -- ECCV 2024},
  pages     = {363--379},
  publisher = {Springer},
  year      = {2024}
}

@inproceedings{yoon2024frag,
  title={FRAG: Frequency Adapting Group for Diffusion Video Editing},
  author={Yoon, Sunjae and Koo, Gwanhyeong and Kim, Geonwoo and Yoo, Chang D.},
  booktitle={Proceedings of the 41st International Conference on Machine Learning},
  series={Proceedings of Machine Learning Research},
  volume={235},
  pages={57315--57330},
  publisher={PMLR},
  year={2024}
}

@article{xu2024instantmesh,
  title={InstantMesh: Efficient 3D Mesh Generation from a Single Image with Sparse-view Large Reconstruction Models},
  author={Xu, Jiale and Cheng, Weihao and Gao, Yiming and Wang, Xintao and Gao, Shenghua and Shan, Ying},
  journal={arXiv preprint arXiv:2404.07191},
  year={2024}
}

@inproceedings{wang2024crm,
  title     = {{CRM}: Single Image to 3D Textured Mesh with Convolutional Reconstruction Model},
  author    = {Wang, Zhengyi and Wang, Yikai and Chen, Yifei and Xiang, Chendong and Chen, Shuo and Yu, Dajiang and Li, Chongxuan and Su, Hang and Zhu, Jun},
  booktitle = {Computer Vision -- ECCV 2024},
  pages     = {57--74},
  publisher = {Springer},
  year      = {2024}
}

\clearpage
\begin{center}
{\Large\bfseries Supplementary Material for SPECSIA:\\
Stylization Dataset for Novel-View Enhancement\\
in Drawing-based 3D Animation\par}
\end{center}
\vspace{1.0em}

\setcounter{figure}{10}
\setcounter{table}{3}
\setcounter{equation}{3}
\makeatletter
\renewcommand{\theHsection}{supp.section.\arabic{section}}
\renewcommand{\theHsubsection}{supp.subsection.\arabic{section}.\arabic{subsection}}
\renewcommand{\theHsubsubsection}{supp.subsubsection.\arabic{section}.\arabic{subsection}.\arabic{subsubsection}}
\renewcommand{\theHfigure}{supp.figure.\arabic{figure}}
\renewcommand{\theHtable}{supp.table.\arabic{table}}
\renewcommand{\theHequation}{supp.equation.\arabic{equation}}
\makeatother
\section*{Overview}
This supplementary document provides additional details and results that were not included in the main paper due to space limits. It also provides the full protocols and extended tables for the additional analyses summarized in the main paper, including clean-target, dataset-split, cross-pipeline, view-sector, user preference, survey-interface details, lightweight-adaptation examples, high-resolution scalability, patch-wise inference analysis, efficiency, and limitation analyses.

\appendix

\section{Mixamo Motion List}\label{sec:supp_mixamo}

\begin{table}
    \centering
    \caption{Mixamo motion list used for evaluation. Motions are grouped by the dominant view tendency.}
    \label{tab:mixamo_motion_list}
    \setlength{\tabcolsep}{3pt}
    \renewcommand{\arraystretch}{1.10}
    \begin{tabular}{p{0.45\linewidth}|p{0.45\linewidth}}
        \toprule
        \textbf{Frontal-view motions} & \textbf{Novel-view motions} \\
        \midrule
        Catwalk Walk Forward & Unarmed Turn Left 90 \\
        Dismissing Gesture & Right Turn W/ Briefcase \\
        Strafe & Stand To Cover \\
        Opening & Harvesting \\
        Button Pushing & Nervously Look Around \\
        Walk To Stop & Ninja Idle \\
        Backwards Rifle Walk & Sword And Shield Turn \\
        Great Sword Strafe & Mutant Punch \\
        Action Idle To Standing Idle & Sword And Shield Attack \\
        Standing Walk Right & Climbing Ladder \\
        \bottomrule
    \end{tabular}
\end{table}

We use a fixed set of Mixamo motions for each test character during evaluation \cite{Mixamo}.
As described in the main paper, we split the motions into frontal-view and novel-view categories.
Frontal-view motions mostly keep the character facing the camera, whereas novel-view motions involve frequent turns or back-facing poses, exposing unseen regions.
This classification is determined by whether the character's face or torso is predominantly oriented toward the camera.
Table \ref{tab:mixamo_motion_list} lists the motion names used in our experiments.

\section{Additional Experimental Results}\label{sec:supp_more_exp}
This section provides additional qualitative and quantitative comparisons on the held-out evaluation setting described in the main paper, including the character-disjoint 3DBiCar test split and hand-drawn Amateur Drawings examples driven by the fixed Mixamo motion set.
The evaluation setting is related to recent efforts on robust image animation and geometry-aware editing \cite{yoon2024tpc, koo2024flexiedit, koo2025flowdrag}.
We sample the hand-drawn characters from the Amateur Drawings dataset \cite{10.1145/3592788}, following the evaluation setting used in DrawingSpinUp \cite{zhou2024drawingspinup} and the Occlusion-robust Stylization Framework \cite{yoon2025occlusion}.
All examples are generated using the same inputs and settings as the main paper, with identical 3D reconstruction and rigging inputs across methods.
Red boxes highlight projection artifacts and the corresponding corrections by our method.

\begin{figure}[p]
\centering
\includegraphics[width=\linewidth]{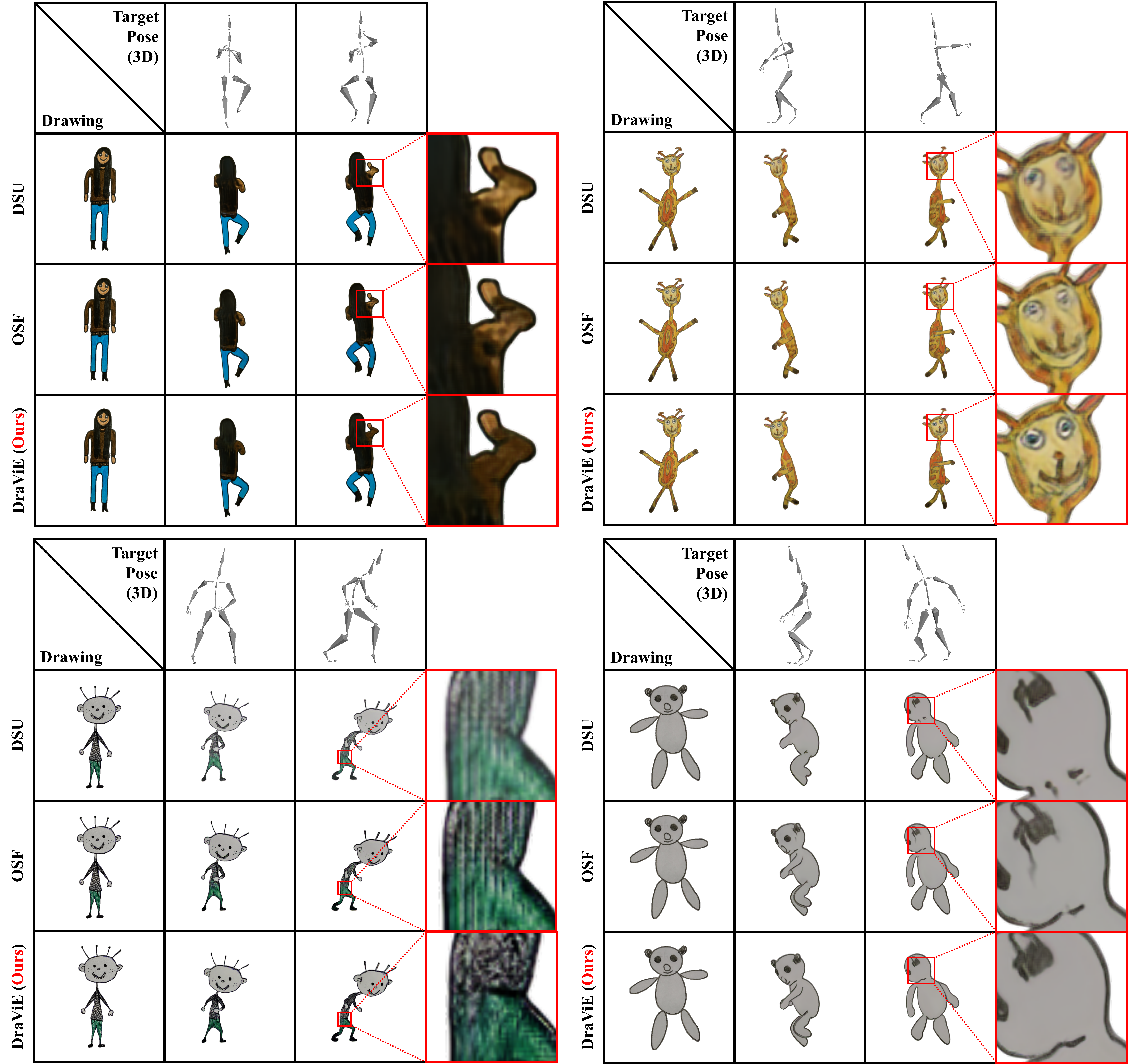}
\caption{Additional qualitative comparisons (set 1) among DSU, OSF, and our method. See the project page for videos.}
\label{fig:supp_qual_1}
\end{figure}

As shown in Fig. \ref{fig:supp_qual_1}, our method improves multiple types of view-dependent artifacts.
(i) It reduces projection artifacts and speckle noise (left column),
(ii) It better preserves character appearance under challenging motions with large viewpoint changes (top-right example),
and (iii) It refines line structures while maintaining the original drawing style (bottom-right example).

\begin{figure}[p]
\centering
\includegraphics[width=\linewidth]{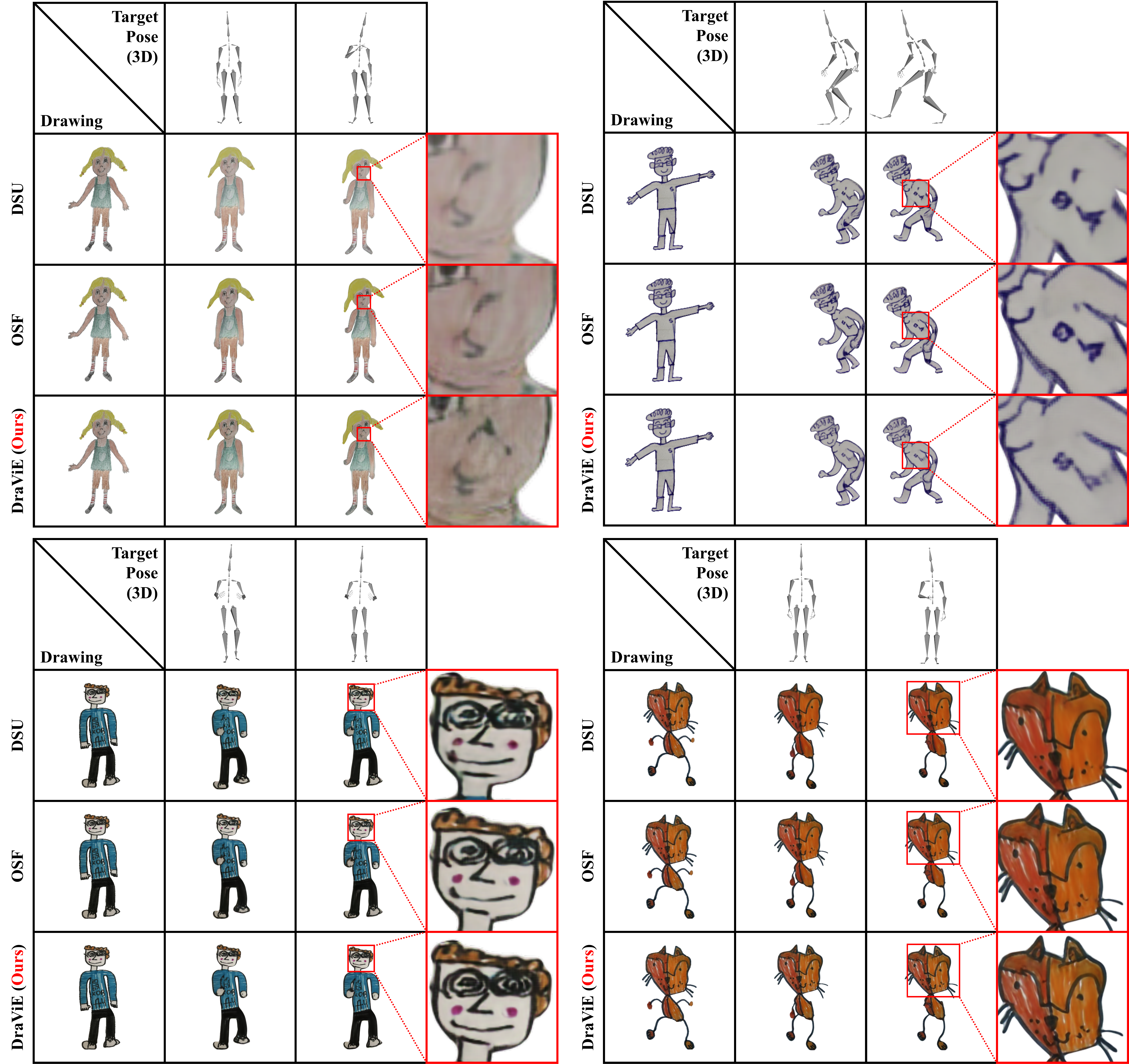}
\caption{Additional qualitative comparisons (set 2). See the project page for videos.}
\label{fig:supp_qual_2}
\end{figure}

As shown in Fig. \ref{fig:supp_qual_2}, our method better preserves fine-grained appearance details under viewpoint changes.
(i) It maintains local attributes such as mouth shape, clothing logos, glasses, and hair color (top row and bottom-left example),
and (ii) It also preserves high-frequency stroke patterns in colored regions (e.g., marker-like textures) without over-smoothing (bottom-right example).

\begin{table}[t]
    \caption{Additional quantitative evaluations on drawing-based 3D animation models. }
    \label{tab:supp_quantitative}
    \centering
    \setlength{\tabcolsep}{5pt}
    \renewcommand{\arraystretch}{1.10}

    \resizebox{\linewidth}{!}{%
        \begin{tabular}{lcccccccc}
            \toprule
            & \multicolumn{4}{c}{\textbf{Frontal-view}} & \multicolumn{4}{c}{\textbf{Novel-view}}\\
            \cmidrule(lr){2-5}\cmidrule(lr){6-9}
            \textbf{Method}
            & \textbf{CLIP}$\uparrow$
            & \textbf{SSIM}$\uparrow$
            & \textbf{LPIPS}$\downarrow$
            & \textbf{FID}$\downarrow$
            & \textbf{CLIP}$\uparrow$
            & \textbf{SSIM}$\uparrow$
            & \textbf{LPIPS}$\downarrow$
            & \textbf{FID}$\downarrow$ \\
            
            \midrule
            DSU \cite{zhou2024drawingspinup} & 0.878 & 0.834 & 0.216 &  211.03 & 0.850 & 0.831 & 0.235 & 267.67 \\
            OSF \cite{yoon2025occlusion} & 0.887 & 0.835 & 0.215 & 206.39 & 0.858 & 0.832 & 0.233 & 249.76 \\
            \rowcolor{green!10} \textbf{DraViE (Ours)} & \textbf{0.888} & \textbf{0.835} & \textbf{0.214} & \textbf{201.51} & \textbf{0.859} & \textbf{0.832} & \textbf{0.233} & \textbf{213.40} \\
            \bottomrule
        \end{tabular}%
    }
\end{table}

Table \ref{tab:supp_quantitative} shows that DraViE generalizes to the Amateur Drawings dataset and improves both fidelity and realism, especially in novel views.

\section{Additional Analyses}\label{sec:supp_cr_additions}

The main paper reports a compact summary of the cross-pipeline and user-preference results. Here, we provide the full supporting analyses, including direct clean-target metrics, descriptor-space backend discrepancy, dataset-split results, view-sector robustness, user preference, survey-interface details, high-resolution scalability, patch-wise inference behavior, efficiency, and extended limitation cases.

\subsection{Cross-pipeline generalization}

\begin{table}
\footnotesize
\renewcommand{\arraystretch}{1.05}
\caption{Descriptor-space backend discrepancy and cross-pipeline evaluation.
Left: descriptor-space discrepancies among raw projected renderings from different upstream reconstruction backends before 2D refinement.
FDD/SW1/C/W denote Fréchet Descriptor Distance, sliced Wasserstein-1 distance, and cross/within discrepancy ratio, respectively.
Right: each triplet reports DSU/OSF/DraViE.}
\label{tab:supp_pipeline}
\centering
\begin{minipage}{0.40\textwidth}
\centering
\setlength{\tabcolsep}{1pt}
\begin{tabular}{lccc}
\toprule
Pair & FDD$\downarrow$ & SW1$\downarrow$ & C/W$\downarrow$\\
\midrule
CRM--IM & 37.67 & 0.868 & 12.89\\
CRM--W3D & 108.43 & 1.503 & 20.51\\
IM--W3D & 63.63 & 1.098 & 17.31\\
\bottomrule
\end{tabular}
\end{minipage}
\hfill
\begin{minipage}{0.55\textwidth}
\centering
\resizebox{\linewidth}{!}{%
\begin{tabular}{lccc}
\toprule
Upstream & CLIP$\uparrow$ & LPIPS$\downarrow$ & FID$\downarrow$\\
\midrule
W3D \cite{long2023wonder3d} & .874/.877/.882 & .231/.230/.225 & 243/236/203\\
IM \cite{xu2024instantmesh} & .831/.835/.840 & .277/.275/.272 & 242/233/205\\
CRM \cite{wang2024crm} & .807/.810/.848 & .272/.271/.267 & 271/266/239\\
\bottomrule
\end{tabular}}
\end{minipage}
\end{table}

\begin{figure}
\centering
\includegraphics[width=\linewidth]{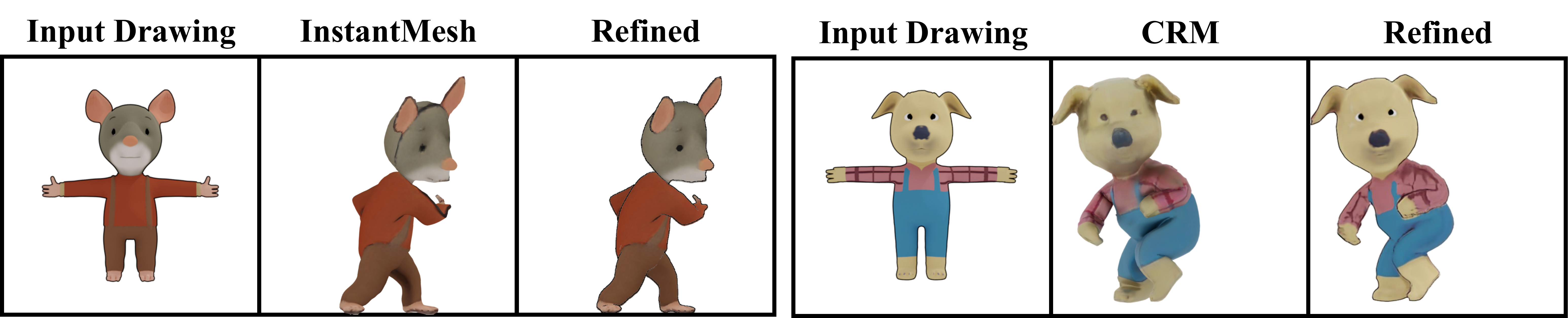}
\caption{Cross-pipeline qualitative examples with InstantMesh \cite{xu2024instantmesh} and CRM \cite{wang2024crm} backends.
The same SPECSIA-pretrained checkpoint and lightweight adaptation protocol are used across backends, without backend-specific retraining or architectural changes.}
\label{fig:supp_cross_pipeline_qual}
\end{figure}

Here, \emph{raw projected renderings} refer to the unrefined 2D color frames produced by each upstream reconstruction backend before applying DSU, OSF, or DraViE.
The left part of Table~\ref{tab:supp_pipeline} quantifies backend discrepancy in a descriptor space built from foreground color histograms, edge statistics, high-frequency residual statistics, and mask area.
FDD denotes a Fréchet Descriptor Distance, computed as an FID-like Gaussian distance in this descriptor space,
SW1 denotes sliced Wasserstein-1 distance over standardized descriptors, and
C/W denotes a cross/within ratio, i.e., the cross-backend discrepancy normalized by the average within-backend discrepancy.
These metrics are diagnostics for backend distribution shifts, not final animation-quality scores.
For cross-pipeline evaluation, we use the same SPECSIA-pretrained checkpoint for all upstream backends and apply the same one-epoch per-character lightweight adaptation protocol when adaptation is used.
We do not perform backend-specific pretraining, architecture changes, loss changes, or retraining.
Thus, this setting evaluates whether the learned correction prior transfers beyond Wonder3D-specific artifacts while keeping the test-time calibration procedure fixed.
As shown in Table~\ref{tab:supp_pipeline}, DraViE consistently improves over DSU/OSF across upstream backends.
Fig.~\ref{fig:supp_cross_pipeline_qual} further shows that the artifact characteristics vary by backend: InstantMesh produces projections that are visually similar to the Wonder3D pipeline, but texture errors often appear near regions where the front and back surfaces meet, resembling pieces of paper pasted together, while CRM produces clay-like surface textures that deviate from the original drawing texture.
Despite these backend-specific artifacts, DraViE reduces projection-level speckles, boundary corruption, and texture inconsistencies.

\subsection{Direct clean-target and dataset-split evaluation}

\begin{table}
\footnotesize
\renewcommand{\arraystretch}{1.05}
\caption{Direct clean-target and dataset-split evaluations. P/S/L/F/C denote PSNR/SSIM/LPIPS/FID/CLIP.}
\label{tab:supp_clean_split}
\centering
\begin{minipage}{0.45\textwidth}
\centering
\begin{tabular}{lccc}
\toprule
Method & P$\uparrow$ & S$\uparrow$ & L$\downarrow$\\
\midrule
Input proj. & 11.785 & .793 & .309\\
DSU & 11.963 & .808 & .263\\
OSF & 11.975 & .810 & .259\\
DraViE & \textbf{12.053} & \textbf{.811} & \textbf{.257}\\
\bottomrule
\end{tabular}
\end{minipage}
\hfill
\begin{minipage}{0.5\textwidth}
\centering
\resizebox{\linewidth}{!}{%
\begin{tabular}{lccc|ccc}
\toprule
& \multicolumn{3}{c|}{3DBiCar} & \multicolumn{3}{c}{Amateur}\\
Method & C$\uparrow$ & L$\downarrow$ & F$\downarrow$ & C$\uparrow$ & L$\downarrow$ & F$\downarrow$\\
\midrule
DSU & .877 & .235 & 319.4 & .871 & .227 & 217.4\\
OSF & .881 & .234 & 274.1 & .872 & .222 & 207.9\\
DraViE & \textbf{.887} & \textbf{.229} & \textbf{223.3} & \textbf{.872} & \textbf{.221} & \textbf{190.5}\\
\bottomrule
\end{tabular}}
\end{minipage}
\end{table}

Table~\ref{tab:supp_clean_split} reports direct PSNR/SSIM/LPIPS against matched clean 3DBiCar renders and separated results based on 3DBiCar/Amateur Drawings.
DraViE improves clean-target fidelity and retains improvements on held-out Amateur Drawings, indicating that the gains are not restricted to the training-domain character assets.

\subsection{View-sector analysis}

\begin{table}
\footnotesize
\caption{View-sector analysis. C/L denote CLIP/LPIPS.}
\label{tab:supp_view_sector}
\centering
\resizebox{0.85\linewidth}{!}{%
\begin{tabular}{lcccccc}
\toprule
& \multicolumn{2}{c}{$0$--$45^\circ$} & \multicolumn{2}{c}{$45$--$90^\circ$} & \multicolumn{2}{c}{$90$--$180^\circ$}\\
Method & C$\uparrow$ & L$\downarrow$ & C$\uparrow$ & L$\downarrow$ & C$\uparrow$ & L$\downarrow$\\
\midrule
DSU & .9037 & .2869 & .8570 & .3216 & .8364 & .3312\\
OSF & .9098 & .2845 & .8635 & .3204 & .8428 & .3307\\
DraViE & \textbf{.9111} & \textbf{.2845} & \textbf{.8635} & \textbf{.3197} & \textbf{.8449} & \textbf{.3299}\\
\bottomrule
\end{tabular}}
\end{table}

Table~\ref{tab:supp_view_sector} reports metrics by yaw sector.
For each sector, we filter animation frames and score only frames whose dominant character orientation falls within the corresponding yaw range.
The $0$--$45^\circ$ sector uses frames from Catwalk Walk Forward, Dismissing Gesture, Strafe, Opening, Button Pushing, Walk To Stop, Backwards Rifle Walk, Great Sword Strafe, Action Idle To Standing Idle, and Standing Walk Right.
The $45$--$90^\circ$ sector uses frames from Unarmed Turn Left 90, Right Turn W/ Briefcase, Harvesting, Nervously Look Around, Sword And Shield Turn, and Mutant Punch.
The $90$--$180^\circ$ sector uses frames from Stand To Cover, Ninja Idle, Sword And Shield Attack, and Climbing Ladder.
DraViE is best or tied-best across sectors, including the more challenging $90$--$180^\circ$ range.
Exact recovery of invisible face or eye details in back views remains ambiguous; the goal is plausible style-consistent artifact reduction conditioned on the projected proxy, mask, and positional hint.

\subsection{User preference study}

\begin{table}
\footnotesize
\caption{Blinded user preference. Preference values are percentages.}
\label{tab:supp_user_pref}
\centering
\begin{tabular}{lccc}
\toprule
Method & Artifact & Style & Overall\\
\midrule
DSU/OSF & 26.7 & 40.0 & 13.3\\
DraViE & \textbf{73.3} & \textbf{60.0} & \textbf{86.7}\\
\bottomrule
\end{tabular}
\end{table}

\begin{figure}
\centering
\includegraphics[width=\linewidth]{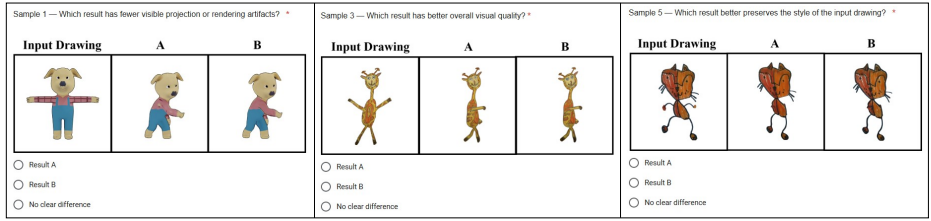}
\caption{Example interface used in the blinded A/B user preference study.
Participants were shown the input drawing and two anonymized results, labeled only as A and B.
They selected Result A, Result B, or no clear difference for questions on artifact reduction, style preservation, and overall visual quality.}
\label{fig:supp_survey_example}
\end{figure}

We ran a 30-person blinded A/B preference study with hidden method names and randomized result order.
Each participant answered five comparison questions, where DraViE was compared against a randomly selected prior method (DSU or OSF).
As shown in Table~\ref{tab:supp_user_pref}, participants preferred DraViE for artifact reduction, style preservation, and overall visual quality.
Fig.~\ref{fig:supp_survey_example} shows representative survey screens; the anonymized A/B interface reduces potential bias in the preference responses.

\subsection{Challenging motion subset}

\begin{table}
\centering
\footnotesize
\caption{Challenging motion subset evaluation.}
\label{tab:supp_dynamic_subset}
\setlength{\tabcolsep}{8pt}
\renewcommand{\arraystretch}{1.05}
\begin{tabular}{lcc}
\toprule
Method & CLIP$\uparrow$ & LPIPS$\downarrow$\\
\midrule
DSU & .847 & .250\\
OSF & .850 & .249\\
DraViE & \textbf{.859} & \textbf{.245}\\
\bottomrule
\end{tabular}
\end{table}

The challenging subset consists of motions with frequent viewpoint changes or back-facing poses: Unarmed Turn Left 90, Right Turn W/ Briefcase, Harvesting, Sword And Shield Turn, Mutant Punch, Stand To Cover, Ninja Idle, Sword And Shield Attack, and Climbing Ladder.
As in the view-sector analysis, we score only frames dominated by the target viewpoint range.
Table~\ref{tab:supp_dynamic_subset} shows that DraViE improves reference consistency on this challenging subset, supporting the temporal and viewpoint-robustness trends reported in the main paper.

\subsection{Scalability and efficiency}

\begin{table}[t]
\centering
\footnotesize
\caption{Resolution and efficiency results. Time denotes per-sample adaptation time.}
\label{tab:supp_efficiency}
\begin{minipage}{0.48\linewidth}
\centering
\resizebox{\linewidth}{!}{%
\begin{tabular}{lcccccc}
\toprule
\multicolumn{7}{c}{$512^2$}\\
Method & FPS & GB & Time & CLIP$\uparrow$ & LPIPS$\downarrow$ & FID$\downarrow$\\
\midrule
DSU & 3.31 & 4.00 & 10.48 & .874 & .231 & 243\\
OSF & 7.20 & 2.59 & 6.45 & .877 & .230 & 236\\
DraViE & 7.00 & 2.98 & 6.05 & \textbf{.882} & \textbf{.226} & \textbf{203}\\
\bottomrule
\end{tabular}}
\end{minipage}
\hfill
\begin{minipage}{0.48\linewidth}
\centering
\resizebox{\linewidth}{!}{%
\begin{tabular}{lcccccc}
\toprule
\multicolumn{7}{c}{$1080^2$}\\
Method & FPS & GB & Time & CLIP$\uparrow$ & LPIPS$\downarrow$ & FID$\downarrow$\\
\midrule
DSU & 0.87 & 15.07 & 47.22 & .831 & .230 & 249\\
OSF & 1.83 & 10.22 & 30.65 & .833 & .229 & 234\\
DraViE & 1.75 & 11.90 & 26.83 & \textbf{.838} & \textbf{.228} & \textbf{204}\\
\bottomrule
\end{tabular}}
\end{minipage}
\end{table}

\begin{figure}[p]
\centering
\includegraphics[width=\linewidth]{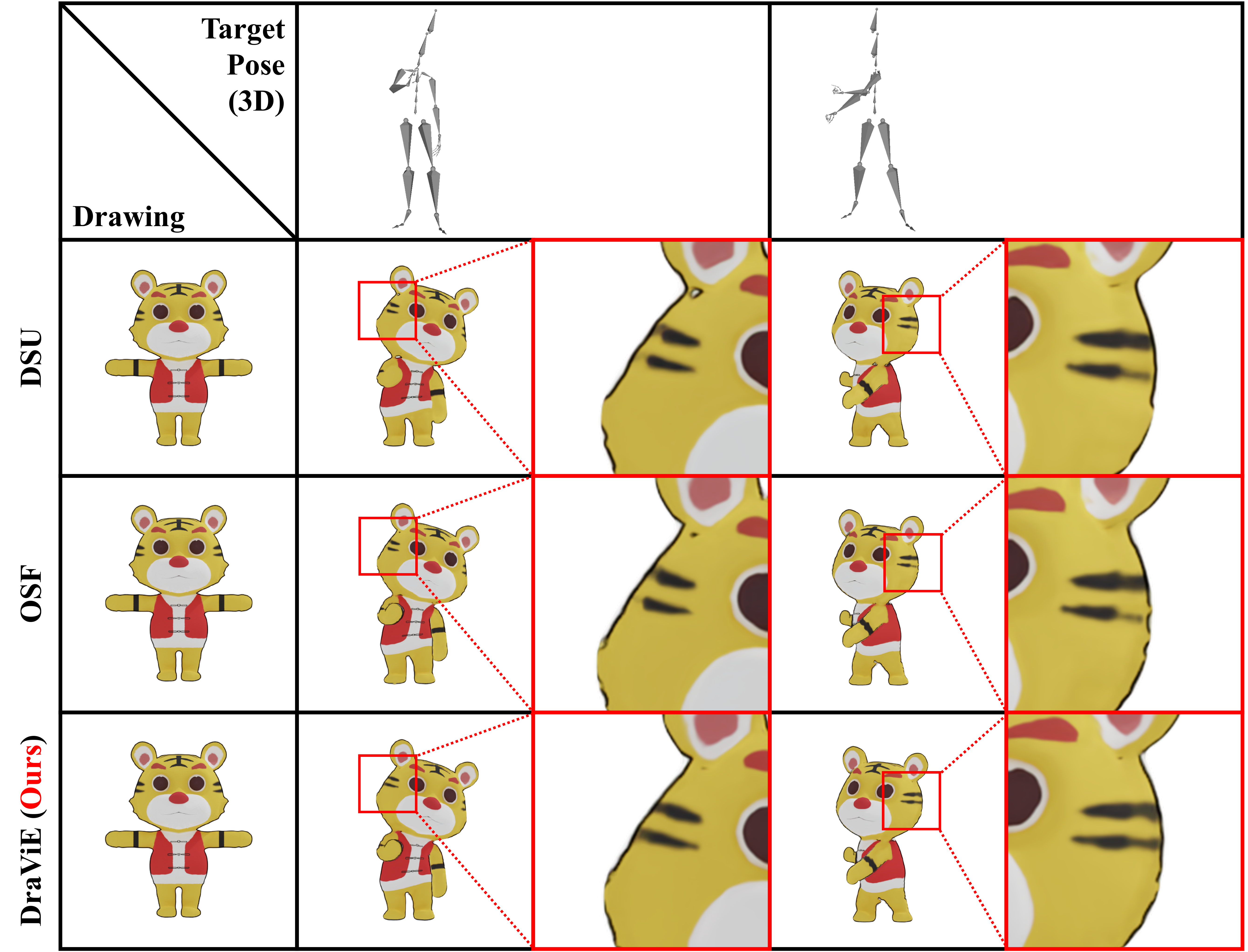}
\caption{High-resolution qualitative examples at $1080^2$.
We show one character under two different target motions, comparing DSU, OSF, and DraViE using the same high-resolution rendering setting.
Zoomed regions highlight local texture and boundary artifacts around the face and body.
DraViE reduces projection-level distortions and produces cleaner local structures while preserving the input drawing style.}
\label{fig:supp_highres_1080}
\end{figure}

\begin{figure}[p]
\centering
\includegraphics[width=\linewidth]{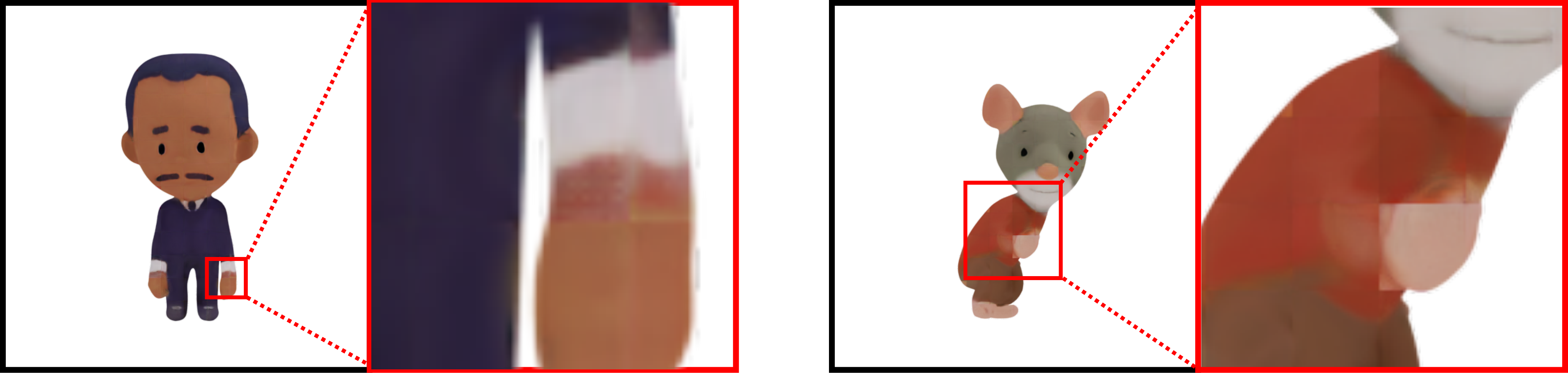}
\caption{Non-overlapping patch-wise inference artifacts.
Although each generated patch can be locally plausible, independently processing and stitching non-overlapping patches can break continuity across patch boundaries.
Visible grid-like seams appear in local color and texture regions, motivating our use of full-frame inference for the reported results.}
\label{fig:supp_patch_inference}
\end{figure}

Table~\ref{tab:supp_efficiency} reports full-frame inference speed, peak VRAM, per-sample adaptation time, and quality metrics at $512^2$ and $1080^2$ resolutions.
DraViE is fully convolutional and runs on full frames at both resolutions.
Fig.~\ref{fig:supp_highres_1080} shows that the same full-frame inference pipeline remains applicable at $1080^2$, reducing local texture corruption and boundary distortions that remain visible in DSU and OSF.
Patch-wise inference can reduce memory, but Fig.~\ref{fig:supp_patch_inference} shows why we do not use non-overlapping patch inference for final results: patch-level outputs may look locally reasonable, yet stitching independently generated patches can produce discontinuous color/texture transitions and grid artifacts.
Overlapping patches can reduce seams, but they add redundant computation and runtime overhead.
Mask alignment is performed offline and takes approximately 17 seconds per view in our implementation.

\subsection{Release and reproducibility}
To support reproducibility, we will release the training/evaluation code, dataset generation scripts, split lists, rendering parameters, metadata, motion lists, and redistributable SPECSIA noisy/clean pairs when permitted by the corresponding third-party licenses.
The detailed license and release policy is described in Section~\ref{sec:supp_ethics}.

\subsection{Rigging controls and non-humanoid inputs}
\begin{figure}
\centering
\includegraphics[width=\linewidth]{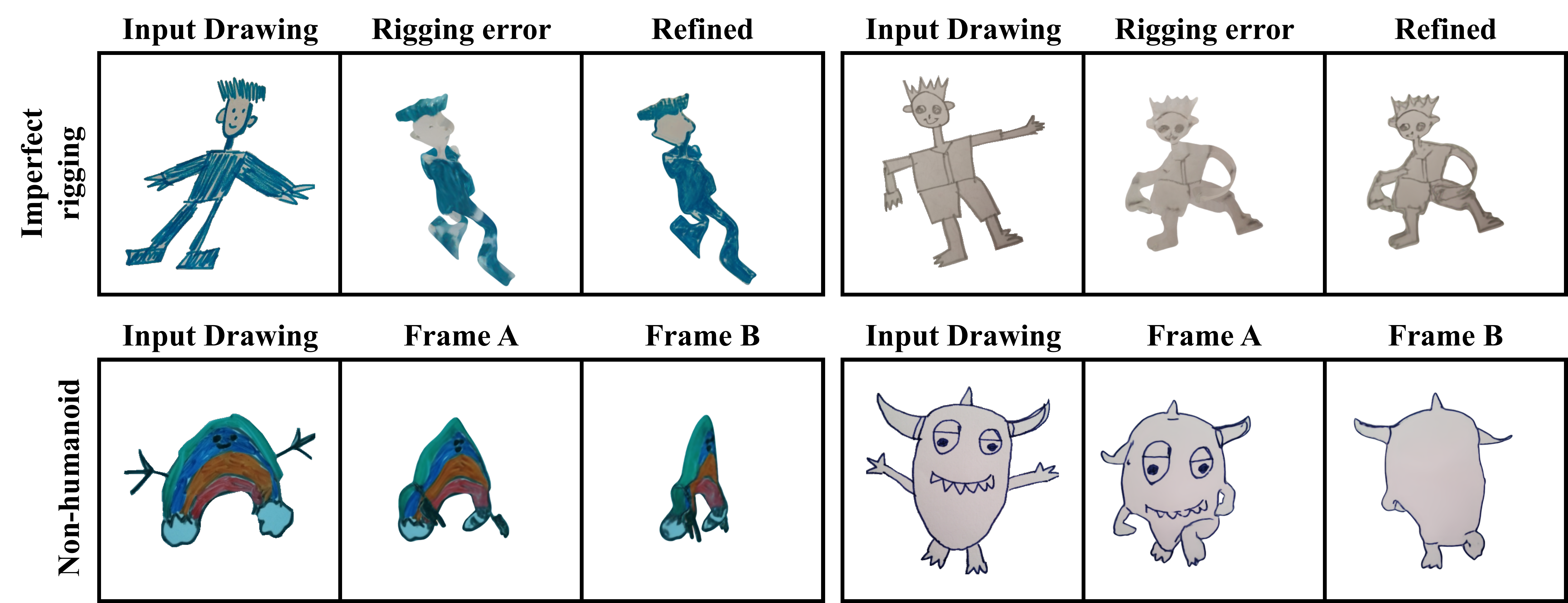}
\caption{Qualitative samples with imperfect rigging and non-humanoid inputs. DraViE improves projection-level artifacts when the proxy remains meaningful, but severe rigging/topology errors remain challenging.}
\label{fig:supp_qual_combined}
\end{figure}

Fig.~\ref{fig:supp_qual_combined} shows representative cases with imperfect rigging and non-humanoid or abstract inputs. DraViE reduces projection-level artifacts when the projected proxy is still structurally meaningful, but it cannot fully repair severe rigging or topology failures. These cases motivate future work on coupling stronger reconstruction, rigging, and temporal refinement with 2D post-correction.

\subsection{Scope and Methodological Limitations}\label{sec:supp_scope_limit}
This work is primarily a dataset-driven projection-refinement study. Unlike test-time calibration or geometry-aware editing methods that directly address compositional alignment or mesh-guided deformation \cite{yoon2024tpc, koo2025flowdrag}, DraViE deliberately keeps the overall drawing-based 3D animation framework and a lightweight 2D refinement backbone close to prior pipelines, so the main contribution lies in scalable paired supervision and data-level correction priors rather than a new 3D reconstruction or temporal-generation architecture. This design choice improves modularity, but it also means that improvements can be moderate when upstream projections are already reasonable.

Because SPECSIA-15K is generated with an offline rendering, reconstruction, and rigging pipeline, it may contain systematic biases from the source assets, the selected camera/view distribution, the reconstruction backend, and the automatic rigging process. 
The clean-target and dataset-split evaluations in Table~\ref{tab:supp_clean_split} and the cross-pipeline evaluation in Table~\ref{tab:supp_pipeline} support that the learned prior transfers beyond the exact training-domain setting, but they do not remove this source of bias entirely. Future dataset construction should include more diverse source assets, reconstruction backends, rigging schemes, and motion categories.

DraViE also refines each frame independently and does not include explicit temporal attention, recurrent memory, or optical-flow constraints. In practice, the shared animated 3D proxy and positional hint provide consistent spatial cues, and the temporal metrics and videos indicate stable behavior under many motions. However, extreme rotations, large self-occlusions, disappearing and reappearing parts, and severe upstream topology errors can still produce flickering or inconsistent details; this is consistent with broader observations in diffusion video editing that high-frequency errors can degrade temporal fidelity \cite{yoon2024frag}. Future work should combine projection refinement with temporal modules and stronger 3D/rigging feedback.

\section{Implementation Details}\label{sec:supp_impl}

\subsection{Network Architecture}\label{sec:supp_arch}
\subsubsection{Backbone. }
Our architecture is based on DrawingSpinUp \cite{zhou2024drawingspinup}.
We adopt a ResNet-based generator with an encoder-decoder structure.
The network consists of three convolutional feature extractors followed by 7 residual blocks and two upsampling blocks.
Specifically, the feature channels are (32, 64, 128) in the encoder and (128, 128, 64) in the decoder with skip connections from encoder stages to decoder stages.
Convolutions use BatchNorm and ReLU/LeakyReLU activations.
For the RIC (Rotation-Invariant Coordinate) convolution \cite{mo2024ric}, standard convolutions are replaced with deformable convolutions using deterministic coordinate offsets.

\subsubsection{Inputs. }
The input is a concatenation of the RGB projection $Z \in \mathbb{R}^{3\times H \times W}$ and auxiliary signals, a binary mask $Z^{mask} \in \mathbb{R}^{1\times H \times W}$ and a 2-channel positional hint $Z^{pos} \in \mathbb{R}^{2\times H \times W}$.
During pretraining, we train on $32\times 32$ patches ($H=W=32$).
We normalize RGB inputs into $[-1, 1]$ with mean/std $(0.5,0.5,0.5)$ and $(0.5,0.5,0.5)$, respectively.
Following DrawingSpinUp \cite{zhou2024drawingspinup}, $Z^{pos}$ encodes the normalized $(x,y)$ coordinates of the character, providing view-independent correspondence cues.
The foreground mask $Z^{mask}$ is obtained from the alpha channel of the RGBA projection. If the alpha channel is missing, we reconstruct it by background thresholding and apply a small dilation to include boundary pixels.

\subsubsection{Outputs. }
The generator outputs $Y\in\mathbb{R}^{3\times H\times W}$ and applies tanh at the last layer.
The discriminator is a PatchGAN-style CNN with instance normalization and $n\_layers=2$,
producing a spatial real/fake score map. Table~\ref{tab:supp_arch} summarizes the generator configuration.

\begin{table}[t]
\centering
\caption{Network architecture summary used in pretraining.}
\label{tab:supp_arch}
\setlength{\tabcolsep}{4pt}
\renewcommand{\arraystretch}{1.10}
    \begin{tabular}{lccc}
    \toprule
    Component & Resolution & Blocks & Channels \\
    \midrule
    Input & $32\times 32$ & -- & $3$(+$1$ mask)(+$2$ pos) \\
    Encoder & $32\rightarrow 16\rightarrow 8$ & 3 conv stages & $32,64,128$ \\
    Latent & $8\times 8$ & 7 ResBlocks & $128$ \\
    Decoder & $8\rightarrow 16\rightarrow 32$ & 2 upsample+conv & $128,128$ \\
    Head & $32\times 32$ & 1 conv block & $64\rightarrow 3$ \\
    Discriminator & $32\times 32$ & PatchGAN ($n{=}2$) & base nf=12 \\
    \bottomrule
    \end{tabular}
\end{table}

\subsection{Training Objective}\label{sec:supp_loss}
We optimize the generator using a weighted sum of reconstruction loss, perceptual loss, and adversarial loss.
Given an input concatenated patch ($Z$ with $Z^{mask}$ and $Z^{pos}$) and the target patch $Y^*$,
the objective is
\begin{align}
    \mathcal{L}
    = \lambda_{recon} \mathcal{L}_{recon}
    + \lambda_{perc} \mathcal{L}_{perc}
    + \lambda_{adv} \mathcal{L}_{adv}.
\end{align}
We set $\lambda_{recon}=4.0$, $\lambda_{perc}=6.0$, and $\lambda_{adv}=0.5$.

\subsubsection{Loss definitions. }
\begin{itemize}
  \item $\mathcal{L}_{recon}$: an L1 loss between the generated patch and the target patch.
  \item $\mathcal{L}_{perc}$: a VGG19 feature loss computed by PerceptualVGG19 \cite{simonyan2015deepconvolutionalnetworkslargescale}.
  We extract features from layers \{0,3,5\} of VGG19.
  We do not apply ImageNet normalization inside the perceptual model, and minimize the mean squared distance between concatenated feature vectors.
  \item $\mathcal{L}_{adv}$: a least-squares GAN loss (LSGAN) implemented with MSELoss.
  The discriminator is trained to classify real targets as 1 and generated outputs as 0, while the generator is trained to fool the discriminator.
\end{itemize}

\subsubsection{Mask usage in objectives. }
In our implementation, all losses are computed on masked regions, not on the full patch.

\subsection{Optimization and Hyperparameters}\label{sec:supp_hparams}
We train with Adam \cite{kingma2015adam} for both the generator and discriminator using the same learning rate $4\times 10^{-4}$ and $(\beta_1,\beta_2)=(0.9,0.999)$ with weight decay $10^{-5}$.
We pretrain for 3 epochs with patch size $32 \times 32$ and batch size 1000.
The pretraining is run with distributed data parallel (DDP) using torchrun, and we save checkpoints only on the main process. Table~\ref{tab:supp_hparams} lists the hyperparameters.

\begin{table}[t]
\centering
\caption{Training hyperparameters for pretraining.}
\label{tab:supp_hparams}
\setlength{\tabcolsep}{6pt}
\renewcommand{\arraystretch}{1.10}
    \begin{tabular}{lc}
    \toprule
    Item & Value \\
    \midrule
    Optimizer (G/D) & Adam / Adam \\
    Learning rate & $4\times 10^{-4}$ \\
    Betas & $(0.9, 0.999)$ \\
    Weight decay & $10^{-5}$ \\
    Batch size & 1000 \\
    Epochs & 3 \\
    Patch size & $32 \times 32$ \\
    Reconstruction loss & L1Loss, weight $4.0$ \\
    Perceptual loss & VGG19 (\{0,3,5\}), weight $6.0$ \\
    Adversarial loss & LSGAN (MSELoss), weight $0.5$ \\
    Discriminator & PatchGAN, $n\_layers{=}2$, base nf=12 \\
    DDP backend & NCCL (torchrun) \\
    \bottomrule
    \end{tabular}
\end{table}

\subsubsection{Patch sampling. }
We construct a set of valid indices from a dilated foreground mask (MaxFilter with kernel size 7), and sample patch centers from an epoch-wise permutation of these indices to minimize repetition within an epoch.
For each sampled center, we crop aligned patches from (i) input projection ($Z$), (ii) positional hint ($Z^{pos}$), (iii) binary mask ($Z^{mask}$), and (iv) target image ($Y^*$).
We also sample an additional random target patch from the same image to support discriminator training.

\subsection{Lightweight Adaptation Details}\label{sec:supp_la}

\subsubsection{What is adapted? }
We perform lightweight adaptation by fine-tuning the full generator network for each target UID (i.e., the character used at inference time).
In this step, we update all generator parameters with standard backpropagation, without changing the network architecture or adding new trainable parameters.
The adaptation is character-specific. For a given UID, we optimize the generator only on that UID's data, which helps preserve the character's appearance and drawing style for projection refinement.

\subsubsection{Budget. }
Lightweight adaptation differs from pretraining in scale and configuration.
For each UID, adaptation uses only a single input image (the reference character image) as training data.
We perform patch-based fine-tuning by sampling many $32\times 32$ patches from this image. This enables a large batch size (1000) even with a single image.
We run a short fine-tuning stage (1 epoch) with the same patch size ($32\times 32$) and batch size (1000), but with a reduced generator learning rate ($4\times 10^{-5}$, $10\times$ lower than pretraining).
We keep the discriminator learning rate at $4\times 10^{-4}$ and use Adam with $(\beta_1,\beta_2)=(0.9,0.999)$ and weight decay $10^{-5}$.
This adaptation stage is executed independently per UID and can be run on a single GPU due to the small per-UID training set, while maintaining the same loss formulation as pretraining. Table~\ref{tab:supp_la} summarizes the per-UID settings.

\begin{table}
\centering
\caption{Lightweight adaptation settings (per-UID fine-tuning).}
\label{tab:supp_la}
\setlength{\tabcolsep}{6pt}
\renewcommand{\arraystretch}{1.10}
    \begin{tabular}{lc}
    \toprule
    Item & Value \\
    \midrule
    Adaptation scope & Per-UID (target character used at inference) \\
    Adapted parameters & Full generator weights \\
    Extra modules & None (no architectural changes) \\
    Epochs & 1 \\
    Patch size & $32 \times 32$ \\
    Batch size & 1000 \\
    Optimizer (G/D) & Adam / Adam \\
    LR (Generator) & $4\times 10^{-5}$ \\
    LR (Discriminator) & $4\times 10^{-4}$ \\
    Betas & $(0.9, 0.999)$ \\
    Weight decay & $10^{-5}$ \\
    Loss weights & $\lambda_{\text{recon}}{=}4.0,\ \lambda_{\text{perc}}{=}6.0,\ \lambda_{\text{adv}}{=}0.5$ \\
    Perceptual backbone & VGG19 layers \{0,3,5\} \\
    Hardware (typical) & Single GPU (due to limited per-UID data) \\
    \bottomrule
    \end{tabular}
\end{table}

\begin{figure}
\centering
\includegraphics[width=\linewidth]{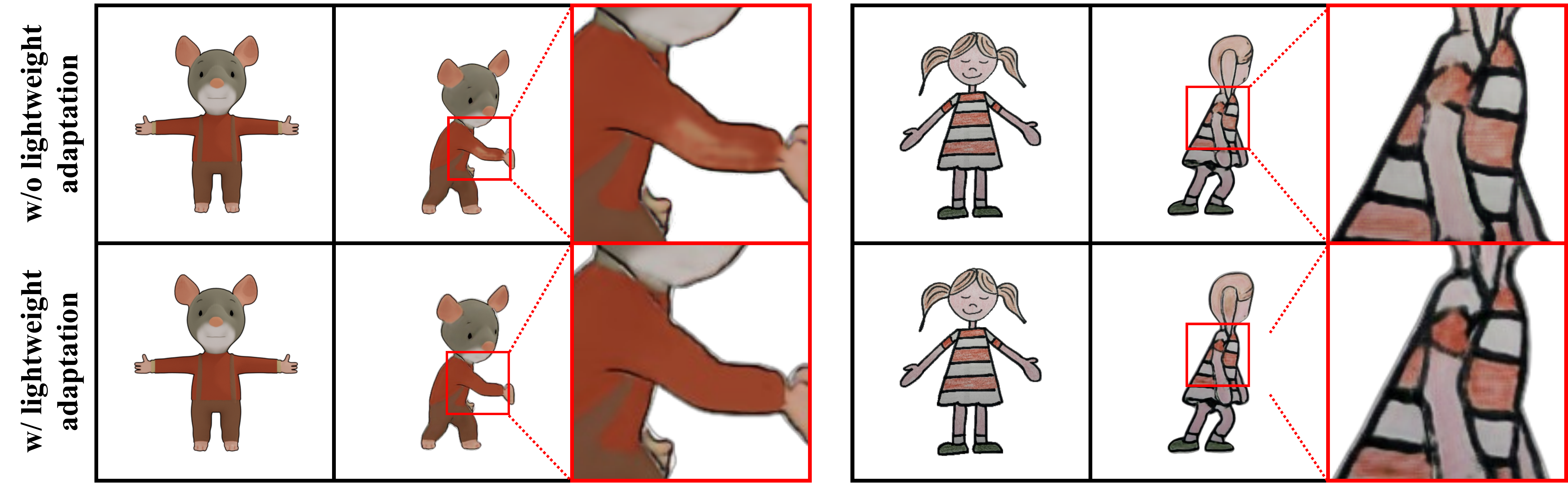}
\caption{Additional qualitative examples of lightweight adaptation under diverse poses.
We compare DraViE without and with lightweight adaptation on two characters and target poses.
Without adaptation, the pre-trained prior can reduce projection artifacts but may over-smooth local colors, strokes, and texture patterns.
Lightweight adaptation better aligns the refined output with the input drawing appearance while retaining the learned projection-correction prior.}
\label{fig:supp_la_examples}
\end{figure}

Fig.~\ref{fig:supp_la_examples} shows additional examples of the lightweight adaptation stage under different poses and character styles.
The adapted model better preserves character-specific appearance cues, such as local color regions, stripe patterns, and stroke thickness, compared with the non-adapted pre-trained model.
These examples complement the lightweight adaptation results in the main paper by illustrating that the adaptation step improves style alignment beyond the single pose shown there.

\section{SPECSIA-15K Dataset Details}\label{sec:supp_data}

\subsection{Data Generation Pipeline}\label{sec:supp_pipeline}

\subsubsection{Source assets. }
We build SPECSIA-15K from the 3DBiCar character collection \cite{luo2023rabit}.
Each character provides a canonical T-pose asset (mesh and texture) and is treated as a unique base identity (UID).
For all experiments in this paper, we use a fixed subset of 1,498 characters and render 10 viewpoints per character, resulting in 14,980 multi-view samples.

\subsubsection{GT renderings. }
We render artifact-free multi-view projections in Blender (v3.6.14) \cite{blender} using a deterministic orthographic setup.
For each character, we (i) import the T-pose meshes, (ii) normalize the scene scale and center the object at the origin, and (iii) render $V = 10$ evenly spaced yaw views in $[0^\circ, 360^\circ)$ (avoiding duplicate $0^\circ$ and $360^\circ$).
We use the Cycles engine with transparent film and save RGBA PNGs at $512 \times 512$ resolution.
To emulate drawing-like silhouettes, we overlay an external contour outline using Blender Freestyle (external contour only), with thickness uniformly sampled from $\{1,2,3,4\}$ pixels and grayscale line color uniformly sampled from $[0,255]$.
We keep the camera trajectory deterministic and disable random pose perturbations to avoid alignment ambiguity.
We use an orthographic camera with \texttt{ortho\_scale=1.35}.
This process produces the GT projections used as supervision targets.

\subsubsection{Packing to the training layout. }
For each rendered RGBA image, we save (i) \texttt{texture.png} (RGBA) and (ii) \texttt{mask.png} by thresholding the alpha channel (\texttt{thr = 1}).
Each view is assigned a unique \texttt{uid\_view} name formatted as \texttt{<char\_id>\_<view\_idx>},
where \texttt{view\_idx} is zero-padded to two digits (e.g., \texttt{0123\_07}).
We store the full list of \texttt{uid\_view} names in a JSON file for training and evaluation.

\subsubsection{Poor projections. }
To construct paired supervision (poor, GT), we generate \emph{poor} projections using a single-view-driven 3D animation pipeline.
Given a source view \texttt{uid\_view} (typically \texttt{\_00}) for each character, we treat it as the single reference input and reconstruct a 3D mesh using Wonder3D \cite{long2023wonder3d}.
We then obtain an animated rest-pose asset via automatic rigging and re-render the mesh under viewpoint changes to synthesize the remaining views.
For each target view, we (i) compute the relative yaw offset between the target and the source view and (ii) render the color projection and a dense positional hint by encoding per-vertex information as vertex colors.
The positional hint stores min-max normalized vertex coordinates in mesh space and is rendered to an RGBA PNG using vertex-color materials, providing view-independent correspondence cues.
Optionally, we perform a mask-alignment step. We render silhouette masks under candidate small translations and pitch/roll offsets and select the best alignment by maximizing the IoU with the GT mask.
We then render the final color/pos using the chosen parameters to reduce structural mismatch between poor projections and GT.
This produces the poor projections that may contain viewpoint-dependent artifacts (e.g., texture mismatch and projection noise), paired with the GT projections rendered directly from the original assets.

\subsection{Splits and Leakage Prevention}\label{sec:supp_split}
We perform character-wise splits by base UID to prevent identity leakage across train/val/test.
After splitting the 1,498 characters, we expand each character into 10 viewpoints, yielding 10 samples per character in each split.
To avoid accidental overlap caused by duplicated renders, we keep the camera path deterministic and ensure that no base UID appears in more than one split. Table~\ref{tab:supp_stats} reports the final split statistics.

\begin{table}
\centering
\caption{Dataset statistics and splits for SPECSIA-15K. Each character contributes 10 viewpoints.}
\label{tab:supp_stats}
\setlength{\tabcolsep}{6pt}
\renewcommand{\arraystretch}{1.10}
    \begin{tabular}{lccc}
    \toprule
    Split & \#Characters & \#Views/Char & Total Images \\
    \midrule
    Train & 1,298 & 10 & 12,980 \\
    Val   & 100 & 10 & 1,000 \\
    Test  & 100 & 10 & 1,000 \\
    \bottomrule
    \end{tabular}
\end{table}

\section{Evaluation Protocols}\label{sec:supp_metrics}

\subsection{Metrics Definitions}\label{sec:supp_metrics_def}
We report three categories of metrics, temporal coherence, fidelity to the reference character, and realism.
All metrics are computed on RGB images. For RGBA outputs we composite onto a solid background when needed.

\subsubsection{Temporal coherence. }
We measure temporal coherence by comparing pairs of frames separated by a temporal stride $s$.
Given a sequence of frames $\{I_t\}_{t=1}^T$, we compute (i) \textbf{CLIP temporal similarity}, the mean cosine similarity between CLIP image features $\mathrm{sim}(\phi(I_t),\phi(I_{t+s}))$ where $\phi(\cdot)$ is L2-normalized, and (ii) \textbf{SSIM temporal}, the mean SSIM between $I_t$ and $I_{t+s}$.
We use OpenCLIP \cite{cherti2023reproducible} with ViT-L/14 (with OpenAI-pretrained weights) to extract $\phi(\cdot)$.
Since our animations are generated by projecting a rigged 3D model, frame-to-frame continuity is already high.
Therefore, we evaluate temporal coherence at a coarser stride and report the results at $s=16$ by default.

\subsubsection{Fidelity (reference consistency). }
For each character-motion-method tuple, we compute reference-based metrics between each generated frame and the reference character texture image (\texttt{texture.png}).
We use \textbf{$LPIPS_{GT}$}, the mean LPIPS \cite{8578166} between each frame and the reference texture (VGG-based LPIPS, resized to $256\times256$), as the primary metric for reference fidelity.
This metric is stride-independent and is computed once per (character, motion, method).

\subsubsection{Realism. }
We report \textbf{FID} \cite{NIPS2017_8a1d6947} between the distribution of reference textures (real set $\mathcal{R}$) and generated frames (fake set $\mathcal{F}$).
We use \texttt{torchmetrics} Frechet Inception Distance with Inception-V3 features (2048-dim).
All images are resized to $299\times299$ and fed as uint8 RGB in batches of 64.
To control the contribution of long videos, we sample a fixed number of frames per video (10 frames per character-motion) and cap the total number of samples per method to 2000.
FID is computed once per method (stride-independent).

\subsection{Sampling Strategy}\label{sec:supp_sampling_notes}
For each character and motion, we evaluate the rendered animation frames produced by each method.
Temporal metrics are computed on frame pairs $(t, t+s)$ for a chosen stride $s$.
In our main results, we use $s=16$ to reflect perceptually meaningful temporal changes while avoiding overly local comparisons between nearly identical consecutive frames.
(For completeness, the evaluation code also supports multiple strides $s \in \{1,2,4,8,16\}$.)
For FID, we sample \textbf{10} frames uniformly at random without replacement from each character-motion video using a deterministic seed, $seed + stable\_hash(character, motion, method)$.
We fix the global random seed (default 0) and enable deterministic computation when specified (\texttt{--deterministic}) to ensure fair and repeatable evaluation.

\subsection{FID Protocol}\label{sec:supp_fid}
We compute FID between (i) \textbf{Real set} (reference textures, \texttt{texture.png}) across evaluated characters and (ii) \textbf{Fake set} (generated frames sampled from each character-motion video, 10 frames per video).
To match sample counts, we randomly subsample both sets to $n=\min(|\mathcal{R}|,|\mathcal{F}|,2000)$ using a fixed seed. Here, $\mathcal{R}$ and $\mathcal{F}$ denote the real and fake sets, respectively.
FID is computed globally per method (not per motion) to obtain a stable estimate of distributional realism.

\begin{figure}
\centering
\includegraphics[width=\linewidth]{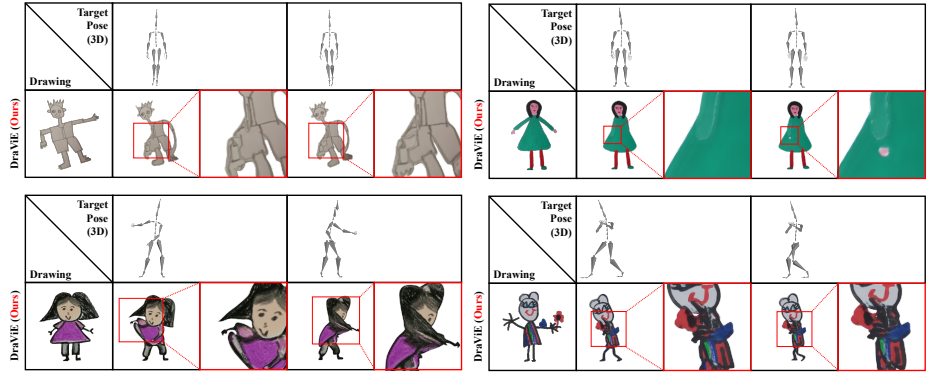}
\caption{Extended failure cases. See the project page for videos.}
\label{fig:supp_fail}
\end{figure}

\section{Extended Failure Cases and Limitations}\label{sec:supp_fail}
As mentioned in the main paper, DraViE is a post-correction module operating on 2D projections and therefore cannot resolve structural errors introduced in upstream 3D reconstruction and rigging.
Fig. \ref{fig:supp_fail} presents representative failure cases.

\subsubsection{Geometric interpenetration. }
In the top row, the arm penetrates the body due to incorrect upstream geometry and pose.
Since the input projection already contains the overlap, DraViE can reduce view-dependent artifacts but cannot recover the correct 3D structure, and the interpenetration remains after refinement.

\subsubsection{Rigging-induced deformation. }
In the bottom-left example, rigging errors cause clothing and hair to unnaturally follow the arm motion.
Such physically incorrect deformations lead to inconsistent projections, and our 2D refinement cannot reliably disentangle hair and clothing boundaries, resulting in appearance inconsistencies.

\subsubsection{Accessory misinterpretation. }
In the bottom-right example, an accessory held in the hand is mistakenly treated as part of the hand by the auto-rigging system.
This produces persistent structural artifacts in the rendered projections, which are also preserved after refinement.

Overall, these cases highlight a fundamental limitation of post-correction. It cannot fix severe upstream 3D failures and should be complemented by more robust reconstruction or rigging when necessary.

\section{Ethics, Licenses, and Release Plan}\label{sec:supp_ethics}

\subsection{Licenses}
SPECSIA-15K is constructed using assets and tools subject to their respective licenses.
We build our dataset from the 3DBiCar character collection \cite{luo2023rabit} and render multi-view projections using Blender \cite{blender}.
Blender is released under the GNU GPL. We use it as an off-the-shelf renderer and do not distribute modified Blender binaries.
For generating poor projections, we use Wonder3D \cite{long2023wonder3d}, which is released under the MIT license, and we follow its license notice and attribution requirements.
For animation and rigging, we use motions from Adobe Mixamo \cite{Mixamo}. According to Adobe's Mixamo licensing FAQ/terms, Mixamo assets may be used in commercial and non-commercial projects without royalty fees. However, redistribution of the downloaded motion files may be restricted.
Accordingly, we do not redistribute any Mixamo motion files, and users should obtain them directly from Mixamo.

\subsection{Intended use}
Our method is intended for research on view-consistent refinement in drawing-based 3D animation pipelines, improving novel-view projection artifacts while preserving the input drawing style.
Potential misuse includes generating higher-quality synthetic animations for deceptive or misleading content, or amplifying stylistic imitation when used with copyrighted or proprietary styles.
We recommend using the method for research and content-creation workflows where the provenance of input assets is legitimate and properly licensed, and we encourage users to disclose synthetic generation when appropriate.

\subsection{Release}
We will release (i) training and evaluation code, (ii) dataset generation scripts, including Blender rendering and packing utilities, (iii) metadata including UID lists, character-disjoint train/validation/test splits, rendering parameters, and the motion list used in evaluation, and (iv) redistributable SPECSIA noisy/clean pairs when permitted by the corresponding third-party licenses.
To respect third-party licenses, we will not redistribute the original 3DBiCar 3D assets \cite{luo2023rabit} or Mixamo motion files \cite{Mixamo}.
If redistribution of any rendered paired projections is restricted, we will provide instructions and scripts to reproduce them from the original sources. This release plan is intended to let users either use the rendered pairs directly, when allowed, or regenerate the same benchmark with the provided split lists and rendering parameters.

\end{document}